\documentclass[twoside,11pt]{article}

\usepackage{jmlr2e}

\usepackage{amsmath}
\usepackage{bbm} 
\usepackage{booktabs}

\newcommand{\Risk}{\mathcal{R}}
\newcommand{\Pm}{\textnormal{P}}
\newcommand{\x}{\mathbf{x}}
\newcommand{\tT}{T}
\newcommand{\Ts}{T^*}
\renewcommand{\mid}{\,|\,}

\usepackage{lastpage}
\jmlrheading{20}{2019}{1-\pageref{LastPage}}{6/18; Revised 7/19}{8/19}{18-424}{H\aa{}vard Kvamme, \O{}rnulf Borgan, and Ida Scheel}

\ShortHeadings{Time-to-Event Prediction with Neural Networks and Cox Regression}{Kvamme, Borgan, and Scheel}
\firstpageno{1}

\begin{document}

\title{Time-to-Event Prediction with Neural Networks\\and Cox Regression}

\author{\name H\aa{}vard Kvamme \email haavakva@math.uio.no \\
       \name \O{}rnulf Borgan \email borgan@math.uio.no \\
       \name Ida Scheel \email idasch@math.uio.no \\
       \addr Department of Mathematics\\
       University of Oslo\\
       P.O. Box 1053 Blindern\\
       0316 Oslo, Norway
       }

\editor{Jon McAuliffe}

\maketitle

\begin{abstract}
    New methods for time-to-event prediction are proposed by extending the Cox proportional hazards model with neural networks. Building on methodology from nested case-control studies, we propose a loss function that scales well to large data sets and enables fitting of both proportional and non-proportional extensions of the Cox model.
    Through simulation studies, the proposed loss function is verified to be a good approximation for the Cox partial log-likelihood.
    The proposed methodology is compared to existing methodologies on real-world data sets and is found to be highly competitive, typically yielding the best performance in terms of Brier score and binomial log-likelihood.
    A python package for the proposed methods is available at \url{https://github.com/havakv/pycox}.
\end{abstract}


\section{Introduction}

In this paper, we consider methodology for time-to-event prediction, a part of survival analysis that reasons about when a future event will occur.
Applications of time-to-event predictions can be found in a variety of settings such as 
survival prediction of cancer patients \citep[e.g.,][]{Vigano2000},
customer churn \citep[e.g.,][]{VANDENPOEL2004196},
credit scoring \citep[e.g.,][]{Dirick2017}, 
and failure times of mechanical systems \citep[e.g.,][]{6879441}.
Arguably, the field of survival analysis has predominantly focused on interpretability, potentially at some cost of predictive accuracy. 
This is perhaps the reason why binary classifiers from machine learning are commonly used in industrial applications where survival methodology is applicable.
However, while the binary classifiers can provide predictions for \textit{one} predetermined duration, one loses the interpretability and flexibility provided by modeling the event probabilities as a function of time. 
Furthermore, in time-to-event data, it is common that some individuals are not followed all the way to their event time, resulting in \textit{censored} times rather than event times.
While binary classifiers typically ignore these observations, one of the main objectives in survival analysis is to account for them.
Hence, in applications with a substantial amount of censoring, the use of survival models tends to be advantageous.

In our work, we propose an approach for combining machine learning methodology with survival models.
We do this by extending the Cox proportional hazards model with neural networks, and further remove the proportionality constraint of the Cox model. Building on methodology from nested case-control studies \citep[e.g.,][]{langholz1996} we are able to do this in a scalable manner. 
The resulting methods have the flexibility of neural networks while modeling event times continuously.
Building on the PyTorch framework~\citep{paszke2017automatic}, we provide  a python package for our methodology, along with all the simulations and data sets presented in this paper.\footnote{Implementations of methods and the data sets are available at \url{https://github.com/havakv/pycox}.}

The paper is organized as follows.
Section~\ref{sec:related_work} contains a summary of related work.
In Section~\ref{sec:methodology}, we review some basic concepts from survival analysis and introduce the Cox proportional hazards model with our extensions.
In Section~\ref{sec:evaluation_criteria} we discuss some evaluation criteria of methods for time-to-event prediction.
In Section~\ref{sec:simulations}, we conduct a simulation study, verifying that the methods we propose behave as expected.
In Section~\ref{sec:experiments} we evaluate our methods on five real-world data sets and compare their performances with existing methodology.
We conclude in Section~\ref{sec:conclusion}.

\section{Related Work}
\label{sec:related_work}

The extension of Cox regression with neural networks was first proposed by \citet{Faraggi1995}, who replaced the linear predictor of the Cox regression model, cf.\ formula
\eqref{eq:cox_prop_haz} below, 
by a one hidden layer multilayer perceptron (MLP). It was, however, found that the model generally failed to outperform regular Cox models \citep{Xiang2000, Sargent2001}.
\citet{DeepSurv} revisited these models in the framework of deep learning and showed that novel networks were able to outperform classical Cox models in terms of the C-index \citep{Harrell1982}.
Our work distinguishes itself from this in the following way:
The method by \citet{DeepSurv}, denoted DeepSurv, is constrained by the proportionality assumption of the Cox model while we propose an extension of the Cox model where proportionality is no longer a restriction.
In this regard, we propose an alternative loss function that scales well for both the proportional and the non-proportional cases.

Similar works based on Cox regression include
SurvivalNet \citep{yousefi2017predicting}, a framework for fitting proportional Cox models with neural networks and Bayesian optimization of the hyperparameters, and
\citet{7822579} and \citet{8100208} which extended the Cox methodology to images. Both \citet{7822579} and \citet{8100208} replace the MLP of DeepSurv with a convolutional neural network and applied these methods to pathological images of lung cancer and to whole slide histopathological images. 

An alternative approach to time-to-event prediction is to discretize the duration and compute the hazard or survival function on this predetermined time grid.
\citet{Luck2017} proposed methods similar to DeepSurv, but with an additional set of discrete outputs for survival predictions and computed an isotonic regression loss over this time grid.
\citet{NeuralMTLR} parameterized a multi-task logistic regression with a neural net that directly computes the survival probabilities on the time grid. 
\citet{deephit} proposed a method, denoted DeepHit, that estimates the probability mass function with a neural net and combine the log-likelihood with a ranking loss; see Appendix~\ref{app:deephit} for details.
Furthermore, the method has the added benefit of being applicable for competing risks. 

The majority of the papers mentioned benchmark their methods against the random survival forests (RSF) by \citet{Ishwaran2008}.
RSF computes a random forest using the log-rank test as the splitting criterion. It computes the cumulative hazards of the leaf nodes and averages them over the ensemble. 
Hence, RSF is a very flexible continuous-time method that is not constrained by the proportionality assumption.

\section{Methodology}
\label{sec:methodology}

In the following, we give a brief review of some concepts in survival analysis. For a more in-depth introduction to the field, see, for example,~\citet{klein2005survival}.

Our objective is to model the event distribution as a continuous function of time.
So with $f(t)$ and $F(t)$ denoting the probability density function and the cumulative distribution function of an event time $\Ts$, we want to model
\begin{align*}
    \Pm(\Ts \leq t) = \int_0^t f(s)\, ds = F(t).
\end{align*}
As alternatives to $F(t)$, it is common to study the \textit{survival function} $S(t)$ and the \textit{hazard rate} $h(t)$.
The survival function is defined as 
\begin{align*}
    S(t) = \Pm(\Ts > t) = 1 - F(t),
\end{align*}
and is commonly used for visualizing event probabilities over time. 
For specifying models, however, it is rather common to use the hazard rate
\begin{align*}
    h(t) = \frac{f(t)}{S(t)}
    = \lim_{\Delta t \rightarrow 0} \frac{1}{\Delta t}\Pm (t \leq \Ts < t + \Delta t \;|\; \Ts \geq t).
\end{align*}
If we have the hazard rate, the survival function can be retrieved through the cumulative hazard,
$H(t) = \int_0^t h(s) ds$, by
\begin{align}
    \label{eq:survival_from_hazard}
    S(t) = \exp[-H(t)]. 
\end{align}
The survival function and the hazard rate therefore provide contrasting views of the same quantities, and it may be useful to study both.

Working with real data, the true event times are typically not known for all individuals. This can occur when the follow-up time for an individual is not long enough for the event to happen, or an individual may leave the study before its termination.
Instead of observing the true event time $\Ts$, we then observe a possibly right-censored event time $\tT = \min\{\Ts, C^*\}$, where $C^*$ is the censoring time.  
In addition, we observe the indicator $D = \mathbbm{1}\{\tT = \Ts\}$ labeling the observed event time $\tT$ as an event or a censored observation.
Now, denoting individuals by $i$, with covariates $\x_i$ and observed duration $\tT_i$,
the likelihood for censored survival times is given by
\begin{align}
    \label{eq:full_likelihood}
    L = \prod_i {f(\tT_i \mid \x_i)}^{D_i} {S(\tT_i \mid \x_i)}^{1 - D_i}
    = \prod_i {h(\tT_i \mid \x_i)}^{D_i} \exp[-H(\tT_i \mid \x_i)].
\end{align}
We will later refer to this as the \textit{full likelihood.}

\subsection{Cox Regression}
\label{sub:cox_regression}

The Cox proportional hazards model~\citep{Cox1972} is one of the most used models in survival analysis. It provides a semi-parametric specification of the hazard rate
\begin{align}
    \label{eq:cox_prop_haz}
    h(t \mid \x) = h_0(t) \exp[g(\x)], \quad\quad  g(\x) = \boldsymbol\beta^T \x,
\end{align}
where $h_0(t)$ is a non-parametric \textit{baseline hazard}, and $\exp[g(\x)]$ is the \textit{relative risk function}. Here, $\x$ is a covariate vector and $\boldsymbol\beta$ is a parameter vector. 
Note that the linear predictor $g(\x) = \boldsymbol\beta^T \x$ does not contain an intercept term (bias weight).
This is because the intercept would simply scale the baseline hazard, and therefore not contribute to the relative risk function,
\begin{align*}
    h_0(t) \exp[g(\x) + b] = h_0(t) \exp[b] \exp[g(\x)] = \tilde h_0(t) \exp[g(\x)].
\end{align*}

The Cox model in~\eqref{eq:cox_prop_haz} is fitted in two steps.
First, the parametric part is fitted by maximizing the \textit{Cox partial likelihood}, which does not depend on the baseline hazard, then the non-parametric baseline hazard is estimated based on the parametric results.
For individual $i$, let $T_i$ denote the possibly censored event time and $\Risk_i$ denote the set of all individuals at risk at time $\tT_i$ (not censored and have not experienced the event before time $\tT_i$). Note that $\Risk_i$ includes individuals with event times at $\tT_i$, so $i$ is part of $\Risk_i$.
The Cox partial likelihood, with Breslow's method for handling tied event times, is given by
\begin{align}
    \label{eq:cox_full_partial_like}
    L_{\text{cox}} = \prod_i {\left(\frac{\exp[g(\x_i)]}{\sum_{j \in \mathcal R_i} \exp[g(\x_j)]} \right)}^{D_i},
\end{align}
and the negative partial log-likelihood can then be used as a loss function
\begin{align}
    \label{eq:cox_loss}
    \textnormal{loss} = \sum_{i} D_i  \log \left (\sum_{j \in \mathcal R_i} \exp[g(\x_j) - g(\x_i)]\right ).
\end{align}
Let $\hat{\boldsymbol \beta}$ be the value of $\boldsymbol\beta$ that maximizes~\eqref{eq:cox_full_partial_like}, or equivalently, minimizes~\eqref{eq:cox_loss}.
Then the cumulative baseline hazard function can be estimated by the \textit{Breslow estimator}
\begin{align}
    \label{eq:breslow_cox}
    \hat H_0(t) &= \sum_{T_i \leq t} \Delta \hat H_0(T_i) \\
    \Delta \hat H_0(T_i) &= \frac{D_i}{\sum_{j \in \mathcal R_i} \exp[\hat g(\x_j)]} \nonumber,
\end{align}
where $\hat g(\x) = {\hat{\boldsymbol\beta}}^T \x$.
If desired, the baseline hazard $h_0(t)$ can be estimated by smoothing the increments, $\Delta \hat H_0(T_i)$, of the Breslow estimate, but the cumulative baseline hazard typically provides the information we are interested in.

\subsection{Cox with SGD}
\label{sub:cox_with_sgd}

The Cox partial likelihood is usually minimized using Newton-Raphson's method.
In our work, we instead want to fit the Cox model with mini-batch stochastic gradient descent (SGD), to better scale to large data sets.
As the loss in~\eqref{eq:cox_loss} sums over risk sets $\Risk_i$, which can be as large as the full data set, it cannot be computed in batches.
Nevertheless, it is possible to do batched iterations by subsampling the data set (to a batch) and restrict the set $\Risk_i$ to only contain individuals in the current batch. 
This scales well for proportional methods such as DeepSurv \citep{DeepSurv}, but would be very computationally expensive for our non-proportional extension presented in Section~\ref{sub:non_proportional_cox}.
Hence, we propose an approximation of the loss that is easily batched.

Intuitively, we can approximate the risk set $\Risk_i$ with a sufficiently large subset $\tilde \Risk_i$, and weight the likelihood accordingly with weights $w_i$,
\begin{align}
    \label{eq:cox_like_samp}
    L = \prod_i {\left(\frac{\exp[g(\x_i)]}
    {w_i  \sum_{j \in \tilde \Risk_i} \exp[g(\x_j)]} \right)}^{D_i}.
\end{align}
The weights should ensure that the weighted sum over the subset $\tilde\Risk_i$ in~\eqref{eq:cox_like_samp} is a reasonable approximation of the full sum over $\Risk_i$ in~\eqref{eq:cox_full_partial_like}.
By choosing a fixed sample size of the sampled risk sets $\tilde \Risk_i$, we can now optimize the objective by batched gradient descent.
The individual $i$ is always included in the sampled risk set $\tilde\Risk_i$ to ensure that each of the products in~\eqref{eq:cox_like_samp} is bounded above by 1.
As the weights $w_i$ do not contribute to the gradients of the logarithm of~\eqref{eq:cox_like_samp} (as can be seen by differentiating with respect to the model parameters), we can simply drop them from the loss function.
Also, in practice we do not compute the loss for $D_i = 0$ as these entries do not contribute to~\eqref{eq:cox_like_samp}. 
Finally, if we average the loss to make it independent of the data set size, we obtain
\begin{align}
    \label{eq:cox_sgd_loss_2}
    \textnormal{loss} = \frac{1}{n}\sum_{i: D_i=1} \log \left (\sum_{j \in \tilde\Risk_i} \exp[g(\x_j) - g(\x_i)]\right ),
\end{align}
where $n$ denotes the number of events in the data set.
In our experiments in Sections~\ref{sec:simulations} and~\ref{sec:experiments}, we find that it is often sufficient to sample only \textit{one} individual $j$ from the risk set, which gives us the loss
\begin{align}
    \label{eq:loss_1}
    \textnormal{loss} = \frac{1}{n}\sum_{i: D_i=1} \log \left ( 1 + \exp[g(\x_j) - g(\x_i)]\right ), \quad j \in \Risk_i \backslash \{i\}.
\end{align}
One benefit of~\eqref{eq:cox_sgd_loss_2} is that it is, in a sense, more interpretable than the negative partial log-likelihood in~\eqref{eq:cox_loss}. 
Due to the sample dependence in the mean partial log-likelihood (MPLL),
i.e., the expression in~\eqref{eq:cox_loss} divided by $n$,
the magnitude of the MPLL is dependent on the size of the risk sets. Hence, for a change of batch size, the mean partial log-likelihood changes. This prohibits a comparison of losses across different batch sizes.
Comparably, the loss in~\eqref{eq:cox_sgd_loss_2} is not affected by the choice of batch size, as the size of $\tilde\Risk_i$ is fixed.
As  a result, we can derive the range of values we expect the loss to be in. 
Using~\eqref{eq:loss_1} as an example, we know that it is typically in the range $(0, \, 0.693]$, as a trivial  $g(\x) = \text{const}$,  gives a loss $= \log(2) \approx 0.693$, and the minimum is obtained by letting $g(\x_i) \to \infty$, $g(\x_j) \to -\infty$, which results in a loss that tends towards 0.

Sampling of the risk sets in Cox's partial likelihood is commonly done in epidemiology and formalized through the nested case-control design, originally suggested by~\citet{Thomas1977}.
In~\eqref{eq:cox_sgd_loss_2}, \textit{case} refers to the $i$'s, while the \textit{controls} are the $j$'s sampled from $\Risk_i \backslash \{i\}$.
\citet{goldstein1992} show that for the Cox partial likelihood, the sampled risk sets produce consistent parameter estimators. While their results do not extend to non-linear models, it is still an indication that the loss function in~\eqref{eq:cox_sgd_loss_2} is reasonable.

Our sampling strategy deviates from that of the nested case-control literature in two ways.
Firstly, we sample a new set of controls for every iteration, instead of keeping control samples fixed.
Secondly, we sample controls with replacement, as this requires less computation than sampling without replacement. 
Note, however, that we typically sample a single control, in which case it does not matter if we sample with or without replacement.

\subsection{Non-Linear Cox}
\label{sub:non_linear_cox}

Having established the simple loss function in~\eqref{eq:cox_sgd_loss_2}, which can be computed  with SGD, the generalization of the relative risk function $\exp[g(\x)]$ is rather straightforward. 
In this paper, we replace the linear predictor $g(\x) = \boldsymbol\beta^T \x$ by a $g(\x)$ parameterized by a neural network.
While our proposed loss function is not a requirement for the adaptation of a neural network~\citep[see, e.g., DeepSurv by ][]{DeepSurv}, it really helps for the further extensions in Section~\ref{sub:non_proportional_cox}. 
Also, it has been found that batched iterations can improve predictive performance~\citep{Kesar2016, Hoffer2017}.

Our generalization of $g(\x)$ leaves the presented theory in Sections~\ref{sub:cox_regression} and~\ref{sub:cox_with_sgd} essentially unchanged, so we do not repeat the likelihoods and loss functions for this model.

We will later refer to the Cox proportional hazards model parameterized with a neural network as Cox-MLP\@. To differentiate between minimizing the negative partial log-likelihood in~\eqref{eq:cox_loss}, as done by DeepSurv, and our case-control approximation in~\eqref{eq:cox_sgd_loss_2}, we will denote the corresponding methods by Cox-MLP (DeepSurv) and Cox-MLP (CC), respectively.

For the non-linear Cox models, the loss does not necessarily have a unique minimizer for $g(\x)$.
Therefore, we add a penalty to the loss function to encourage $g(\x)$ to not deviate too far from zero
\begin{align}
    \label{eq:cox_penalty}
    \text{penalty} = \lambda \sum_{i: D_i=1} \sum_{j \in \tilde\Risk_i} |g(\x_j)|.
\end{align}
Here $\lambda$ is a tuning parameter, and note that $i$ is included in $\tilde\Risk_i$.

\subsection{Non-Proportional Cox-Time}
\label{sub:non_proportional_cox}

The proportionality assumption of the Cox model can be rather restrictive, and parameterizing the relative risk function with a neural net does not affect this constraint.
Approaches for circumventing this restriction are typically based on grouping the data based on a categorical covariate and applying a stratified version of the Cox model \citep[chap.~9]{klein2005survival}.
We propose a parametric approach that does not require stratification.
Continuing with the semi-parametric form of the Cox model, we now let the relative risk function depend on time, 
\begin{align}
    \label{eq:cox_time_model}
    h(t \mid \x) = h_0(t) \exp[g(t, \x)].
\end{align}
In practice, we let $g(t, \x)$ handle the time as a regular covariate, which enables $g(t, \x)$ to model interactions between time and the other covariates.
This is similar to the approach taken in classical survival analysis, where the non-proportional effect of a covariate $x$ may be modeled by including time-dependent covariates like $x\cdot t$ and $x \cdot \log t$.

The model~\eqref{eq:cox_time_model} is no longer a proportional hazards model. However, it is still a relative risk model with the same partial likelihood as previously, only now with an additional covariate. Following the approach from Section~\ref{sub:cox_with_sgd}, we have the loss function
\begin{align}
    \label{eq:loss_cox_time_first}
    \textnormal{loss} = \frac{1}{n} \sum_{i: D_i=1} \log \left (\sum_{j \in \tilde \Risk_i} \exp[g(\tT_i, \x_j) - g(\tT_i,\x_i)]\right ),
\end{align}
and we include the penalty in~\eqref{eq:cox_penalty}, with $g(\tT_i, \x_j)$ replacing $g(\x_j)$.
We will later refer to models fitted by~\eqref{eq:loss_cox_time_first} as \textit{Cox-Time}.

Note that the loss has the same $\tT_i$ for both $\x_i$ and the $\x_j$'s.
Consequently, if we had used the full risk set $\Risk_i$ instead of the subset $\tilde \Risk_i$, as is the case for the loss in~\eqref{eq:cox_loss}, the loss would become very computationally expensive.
In fact, for the full risk set, the time complexity of the loss would be $O(n\cdot |\Risk_i|) = O(n^2)$,
where $|\Risk_i|$ denotes the size of the risk set.
But for~\eqref{eq:loss_cox_time_first} we get $O(n\cdot |\tilde\Risk_i|) = O(n)$, as $|\tilde\Risk_i|$ is fixed and small.
In the proportional case, to compute the loss in~\eqref{eq:cox_loss} one only needs to compute $g(\x_j)$ once (per iteration) for each $j$, and reuse that value in all other risk sets. This ensures the linear time complexity for the classical Cox models.

We can find the Breslow estimate for the cumulative baseline hazard $H_0(t)$ using~\eqref{eq:breslow_cox} with $\hat g(\x_j)$ replaced by $\hat g(\tT_i, \x_j)$.
Note that we still need the non-parametric baseline, as $g(t, \x)$ is restricted to model interactions between the time and the covariates.
To see this, consider $g(t, \x) = a(t, \x) + b(t)$, and observe that $b(t)$ cancels out in the loss.

\subsection{Prediction}
\label{sub:prediction}

We can obtain predictions from the relative risk models by estimating the survival function in~\eqref{eq:survival_from_hazard},
$\hat S(t \mid \x) = \exp[- \hat H(t \mid \x)]$.
For the proportional hazards models, the relative risk function does not depend on time, enabling us to integrate only over the baseline hazard and compute the relative risk separately,
\begin{align*}
    H(t \mid \x) = \int_0^t h_0(s) \exp[g(\x)] \, ds = H_0(t) \exp[g(\x)].
\end{align*}
By first estimating $H_0(t)$ on the training data with~\eqref{eq:breslow_cox}, we only need to compute $\exp[g(\x)]$ to obtain predictions.
Computation of the estimate $\hat H_0(t)$ requires a single pass over the whole training set, in addition to sorting the training set by time.

In the case of models with non-proportional hazards, such as models fitted by Cox-Time in Section~\ref{sub:non_proportional_cox}, predictions are much more computationally expensive.
As the relative risk is time-dependent, we now need to integrate over both the baseline hazard and $g(t, \x)$,
\begin{align*}
    H(t \mid \x)
    = \int_0^t h_0(s) \exp[g(s, \x)] \, ds.
\end{align*}
In practice, we estimate the cumulative hazards by
\begin{align*}
    \hat H(t \mid \x)
    =  \sum_{T_i \leq t} \Delta \hat H_0(T_i) \exp[\hat g(T_i, \x)],
\end{align*}
where $\Delta \hat H_0(\tT_i)$ is an increment of the Breslow estimate and $\hat g(\tT_i, \x)$ is the estimate of $g(\tT_i, \x)$ obtained from the neural network.
This is clearly rather computationally expensive as we need to compute $\hat g(T_i, \x)$ for all distinct event times $T_i \leq t$.
Furthermore, for continuous-time data, computation of the cumulative baseline hazard through the Breslow estimate,
\begin{align}
    \label{eq:breslow_np}
    \Delta \hat H_0(T_i) = \frac{D_i}{\sum_{j \in \Risk_i} \exp[\hat g(T_i, \x_j)]},
\end{align}
scales quadratically.

To alleviate the computational cost, one can compute the cumulative hazards over a reduced number of distinct time points.
Hence, Cox-Time is trained on continuous-time data but produces discrete-time predictions, with the benefit of the discretization happening after the network is fitted.
In practice, we perform this discretization by computing the baseline on a random subset of the training data and subsequently control the resolution of the time grid through the sample size.

\section{Evaluation Criteria}
\label{sec:evaluation_criteria}

Metrics for evaluating the performance of methods for time-to-event prediction should account for the censored individuals.
In the following, we describe the metrics used in the experimental sections of this paper.

\subsection{Concordance Index}
\label{sub:c-index}

In survival analysis, the concordance index, or C-index \citep{Harrell1982}, is arguably one of the most commonly applied discriminative evaluation metrics. 
This is likely a result of its interpretability, as it has a close relationship to classification accuracy~\citep{Ishwaran2008} and ROC AUC \citep{Heagerty2005}.
In short, the C-index estimates the probability that, for a random pair of individuals, the predicted survival times of the two individuals have the same ordering as their true survival times. 
See \citet{Ishwaran2008} for a detailed description.

As the C-index only depends on the ordering of the predictions, it is very useful for evaluating proportional hazards models. This is because the ordering of proportional hazards models does not change over time, which enables us to use the relative risk function instead of a metric for predicted survival time.
It is, however, not obvious how the C-index should be applied for non-proportional hazards models \citep{Gerds2012, Ishwaran2008}.
We will use a metric based on the time-dependent C-index by \citet{Ctd}, which estimates the probability that observations $i$ and $j$ are concordant given that they are comparable,
\begin{align}
    \label{eq:ctd}
    C^{\text{td}} = \Pm\{\hat S(\tT_i \mid \x_i) < \hat S(\tT_i \mid \x_j) \,\mid\, \tT_i < \tT_j,\, D_i = 1\}.
\end{align}
However, to account for tied event times and survival estimates, we make the modifications listed by
\citet[Section 5.1, step 3]{Ishwaran2008}.
This is to ensure that predictions independent of $\x$, $\hat S(t \mid \x) = \hat S(t)$, yields $C^\text{td} = 0.5$ for unique event times.
Note that for proportional hazards models, our metric is equivalent to the regular C-index.

\subsection{Brier Score}
The Brier score (BS) for binary classification is a metric of both discrimination and calibration of a model's estimates.
In short, for $N$ binary labels $y_i \in \{0, 1\}$ with probabilities $p_i$ of $y_i=1$, the BS is the mean squared error of the probability estimates $\hat p_i$, i.e., $\text{BS} = \frac{1}{N}\sum_i {(y_i - \hat p_i)}^2$.
To get binary outcomes from time-to-event data, we choose a fixed time $t$ and label data according to whether or not an individual's event time is shorter or longer than $t$.
\citet{Graf1999} generalize the Brier score to account for censoring by weighting the scores by the inverse censoring distribution,
\begin{align}
    \label{eq:brier_score}
    \text{BS}(t) =  \frac{1}{N} \sum_{i = 1}^N 
    \left [ 
    \frac{{\hat S(t \mid \x_i)}^2\, \mathbbm{1}\{\tT_i \leq t, D_i = 1\}}{\hat G(\tT_i)}
    +
    \frac{{(1 - \hat S(t \mid \x_i))}^2\, \mathbbm{1}\{\tT_i > t\}}{\hat G(t)}
    \right ].
\end{align}
Here $N$ is the number of observations, $\hat G(t)$ is the Kaplan-Meier estimate of the censoring survival function, $\Pm(C^* > t)$, and it is assumed that the censoring times and survival times are independent.

The BS can be extended from a single duration $t$ to an interval by computing the integrated Brier score
\begin{align}
    \label{eq:ibs}
    \text{IBS} = \frac{1}{t_2 - t_1} \int_{t_1}^{t_2} \text{BS}(s)\, ds.
\end{align}
In practice, we approximate this integral by numerical integration, and we let the time span be the duration of the test set. In our experiments we found that using 100 grid points were sufficient to obtain stable scores.

\subsection{Binomial Log-Likelihood}
The mean binomial log-likelihood is a commonly used metric in binary classification that measures both discrimination and calibration of the estimates of a method. Using the same inverse censoring weighting as for the Brier score, we can apply this metric to censored duration time data,
\begin{align}
    \label{eq:bll}
    \text{BLL}(t) = \frac{1}{N} 
    \sum_{i = 1}^N 
    \left [ 
    \frac{\log[1 - \hat S(t \mid \x_i)]\, \mathbbm{1}\{\tT_i \leq t, D_i = 1\}}{\hat G(\tT_i)}
    +
    \frac{\log [\hat S(t \mid \x_i)]\, \mathbbm{1}\{\tT_i > t\}}{\hat G(t)}
    \right ].
\end{align}
The binomial log-likelihood can also be integrated in the same manner as~\eqref{eq:ibs}
\begin{align*}
    \text{IBLL} = \frac{1}{t_2 - t_1} \int_{t_1}^{t_2} \text{BLL}(s)\, ds.
\end{align*}

\section{Simulations}
\label{sec:simulations}

To empirically investigate our proposed methodology, we conducted a series of simulations.
These experiments are by no means exhaustive but are rather meant to verify that the methods behave as expected.
In the following, we let \textit{classical Cox} refer to a Cox regression with $g(\x) = \boldsymbol \beta^T \x$ obtained with the Lifelines python package \citep{cameron_davidson_pilon_2018_1210821}.
For experimental details exempt from the main article, we refer the reader to Appendix~\ref{app:simulation_details}.

We first investigate the behavior of our proposed loss~\eqref{eq:cox_sgd_loss_2}. In particular,
we want to examine the impact the number of sampled controls has on the fitted models, in addition to how well the results from using our loss agree with those from using the Cox partial log-likelihood, i.e., the loss~\eqref{eq:cox_loss}.
To this end, we simulated survival times from a proportional hazards model
\begin{align}
    \label{eq:sim_sim_cox_with_sgd}
    h(t \mid \x) &=  h_0(t) \exp[g(\x)], \nonumber \\
    g(\x) &= \boldsymbol \beta^T \x,
\end{align}
with constant baseline hazard $h_0(t) = 0.1$, and $\boldsymbol \beta^T = [0.44, 0.66, 0.88]$.
The covariates were sampled uniformly from $[-1,\, 1]$.
We drew censoring times independent of the covariates with constant hazard $c(t) =  \frac{1}{30}$,
and, in addition, we censored all individuals that were still under observation at time 30.
This resulted in approximately 30 \% censored individuals.

\begin{figure}[t]
    \begin{center}
        \vspace{5mm}
        \includegraphics[scale=0.45]{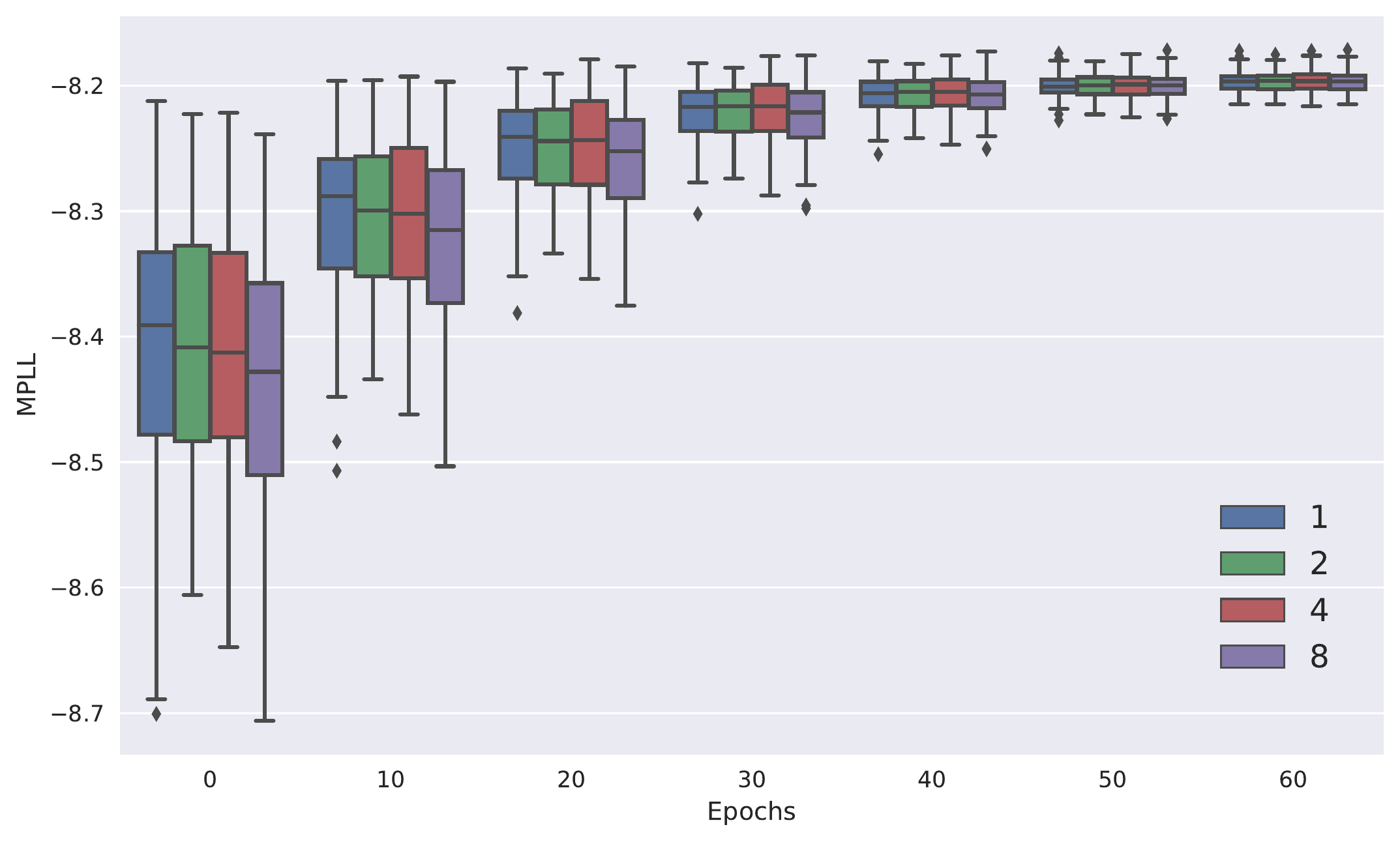}
    \end{center}
    \vspace{-7mm}
    \caption{Box plots giving the mean partial log-likelihood (MPLL) of the test sets for different training epochs. The colors show how many controls were sampled during training (in addition to the case).}\label{fig:cox_control_curves}
\end{figure}

We sampled 10,000 individuals for training and 10,000 for testing, and fitted our Cox model by SGD as described in Section~\ref{sub:cox_with_sgd}. 
This method will be referred to as \textit{Cox-SGD}.
Four  models were fitted by sampling 1, 2, 4, and 8 controls (in addition to the case).
The whole experiment was repeated 100 times, and the mean partial log-likelihood (MPLL) of the test sets are visualized in Figure~\ref{fig:cox_control_curves}. 
The figure indicates that the number of sampled controls does not affect the rate of convergence, but we note that the computational complexity increases with the number of sampled controls.

The experiment was repeated with training sets of 1,000 individuals to verify that the results were not simply due to the size of the training data. 
The same patterns were found in this setting, so the figure is exempted from the paper.

\begin{figure}[t]
    \begin{center}
        \vspace{5mm}
        \includegraphics[scale=0.48]{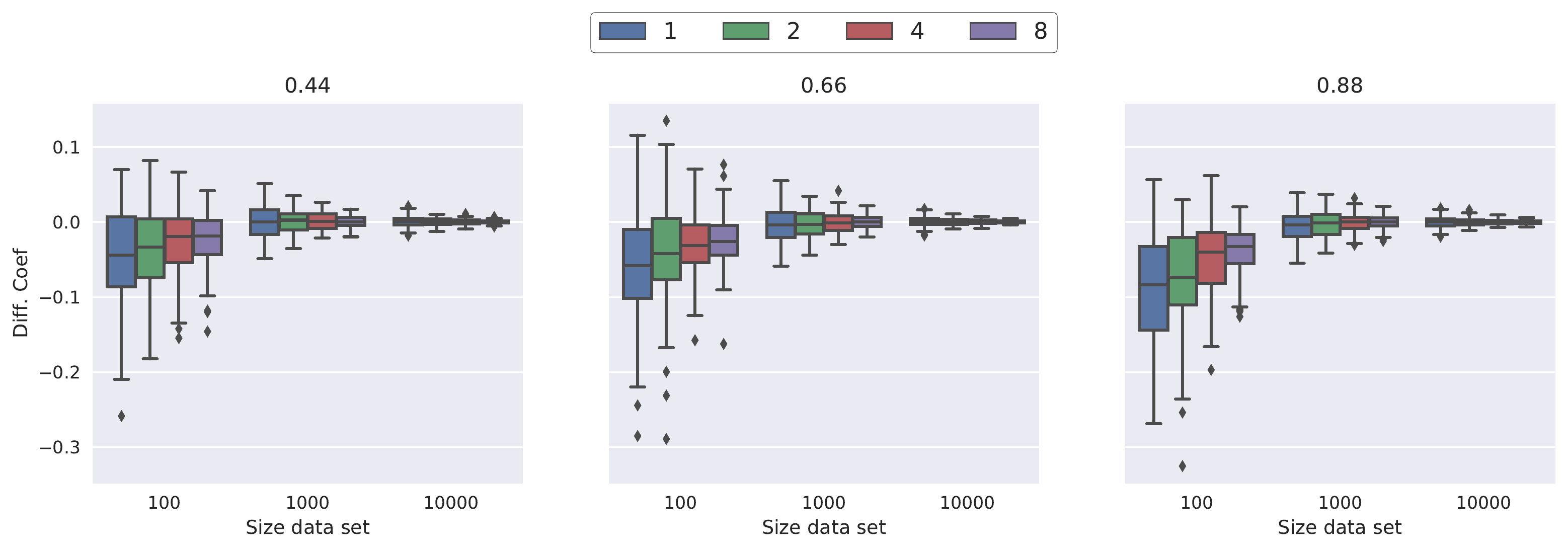}
    \end{center}
    \vspace{-7mm}
    \caption{Differences between Cox-SGD and Classical Cox parameter estimates for different data set sizes. The numbers above the plots give the value of the true coefficient. Each box plot is based on 100 observations. The legend above the plots states the number of sampled controls for Cox-SGD.}\label{fig:diff_coef}
\end{figure}

Next, we compare the parameter estimates obtained from our proposed loss (Cox-SGD) with the estimates obtained with classical Cox regression.
For data sets of sizes 100, 1,000, and 10,000, we fitted models to 100 sampled data sets.
The differences between the Cox-SGD parameter estimates and the classical Cox estimates are displayed in Figure~\ref{fig:diff_coef}, where the legend above the plots gives the number of controls sampled for the Cox-SGD method.
For the data sets of size 100, we observe that the Cox-SGD estimates seem to be slightly smaller than the Cox estimates, and this difference is larger for fewer sampled controls.
However, as the data sets increase in size, the estimates for the two methods agree well.

\begin{figure}[tbp]
    \begin{center}
        \vspace{5mm}
        \includegraphics[scale=0.48]{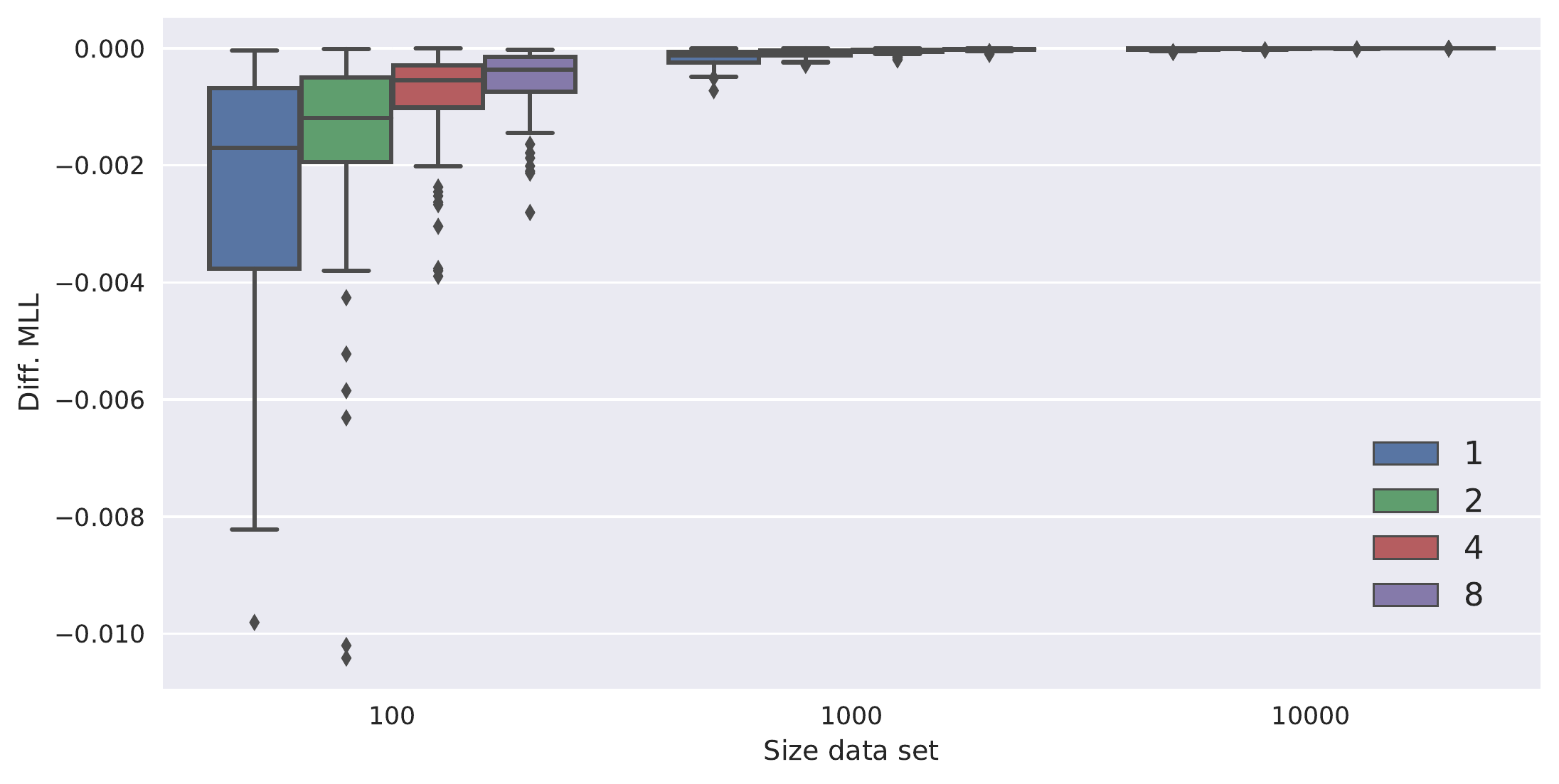}
    \end{center}
    \vspace{-7mm}
    \caption{Differences in mean log-likelihood between Cox-SGD and classical Cox using~\eqref{eq:full_likelihood}.
    The figure gives results for different data set sizes, and the legend gives the number of sampled controls. Each box plot is based on 100 observations.}\label{fig:diff_mll}
\end{figure}

Finally, we want to compare the likelihoods obtained by the two methods. As the mean partial log-likelihood depends on data set size, it is really not comparable across data sets.
We will therefore instead use the full likelihood~\eqref{eq:full_likelihood},
which also depends on the baseline hazard, to compare the methods.
Note that the partial likelihood may be interpreted as a profile likelihood, so the full likelihood and the partial likelihood are closely related
\citep[see, e.g.,][p.~258]{klein2005survival}.
We obtain cumulative baseline hazard estimates by~\eqref{eq:breslow_cox}, and baseline hazard estimates by numerical differentiation of the cumulative baseline hazard estimates.
We report the mean log-likelihood (MLL) of~\eqref{eq:full_likelihood} to compare results for different data set sizes.
In Figure~\ref{fig:diff_mll}, we show the difference in the MLL between the Cox-SGD method and the classical method for Cox regression (for each sampled data set). We used the training MLL as we are here only interested in the losses' abilities to optimize the objective, and not the generalization to a test set.
From the figure, we observe that, for smaller data sets, more sampled controls in Cox-SGD seems to give likelihood estimates closer to those of a classical Cox regression.
As the data sets increase in size, the MLL of the Cox-SGD seems to converge to that of the classical Cox, regardless of control sample size.
The MLL's of the classical Cox regression is approximately -2.2, meaning that even for the smallest data sets the differences in MLL for 1 sampled control is around $0.1$ \%.
Hence, it seems that our loss~\eqref{eq:cox_sgd_loss_2} approximates the negative partial log-likelihood rather well.
Furthermore, while a higher number of sampled controls can give lower training error, the effect of the number of sampled controls decreases with the size of the data sets.  

In the further experiments, we will for simplicity only use one sampled control, as this was found sufficient.
Furthermore, in Appendix~\ref{app:simulations} we have included simulations for the methods using neural networks for non-linear and non-proportional models, to verify that our proposed methods behave as expected.
In summary, these simulations verify that for observations drawn from a proportional hazards model with a non-linear relative risk function, Cox models parameterized by neural networks provide better estimates of the partial log-likelihood than classical Cox models.
Further, by drawing observations from a non-proportional relative risk model, the simulations verify that our Cox-Time method (Section~\ref{sub:non_proportional_cox}) is able to obtain better estimates of the survival functions than methods which assume proportional hazards.

\section{Experiments}
\label{sec:experiments}

In the following, we will evaluate our proposed methods on real data sets and compare their performance to existing methods from the literature.
In total, we use five data sets. One large data set for a more in-depth analysis (see Section~\ref{sec:kkbox_churn_prediction}), and four smaller data sets commonly used in the survival analysis literature.

\subsection{Four Common Survival Data Sets}
\label{sub:five_small_data_sets}

We base this experimental section on the data sets provided by \citet{DeepSurv}, as they are made available through the DeepSurv python package, and need no further preprocessing.
The data sets include the Study to Understand Prognoses Preferences Outcomes and Risks of Treatment (SUPPORT), the Molecular Taxonomy of Breast Cancer International Consortium (METABRIC), and the Rotterdam tumor bank and German Breast Cancer Study Group (Rot.\ \& GBSG).
\citet{DeepSurv} also provide the Worcester Heart Attack Study (WHAS) data set.
However, their version of the data set is actually a case-control data set, meaning it contains multiple replications of some individuals, something the authors seem to have overlooked.
We replace it with the Assay Of Serum Free Light Chain (FLCHAIN) made available in the survival packages of R~\citep{survival-package}.
For FLCHAIN, we remove individuals with missing values. Further, we remove the ``chapter'' covariate, which gives the cause of death.
Table~\ref{tab:dataset_statistics} provides a summary of the data sets.
For a more detailed description, we refer to the original sources \citep{survival-package, DeepSurv}.

\begin{table}[t]
    \centering
    \begin{tabular}{lrrrr}
        \toprule
        Data set &      Size & Covariates & Unique Durations & Prop. Censored \\
        \midrule
        SUPPORT       &  8,873 &   14 &         1,714 &          0.32 \\
        METABRIC      &  1,904 &    9 &         1,686 &          0.42 \\
        Rot.\ \& GBSG &  2,232 &    7 &         1,230 &          0.43 \\
        FLCHAIN       &  6,524 &    8 &         2,715 &          0.70 \\
        \bottomrule
    \end{tabular}
    \caption{Summary of the four data sets used in the experiments in Section~\ref{sub:five_small_data_sets}.}\label{tab:dataset_statistics}
\end{table}

\subsubsection{Methods and Hyperparameter Tuning}
\label{sub:methods}

In Sections~\ref{sub:non_linear_cox} and~\ref{sub:non_proportional_cox} we presented two new survival methods based on case-control sampling and neural networks: a proportional Cox method and a non-proportional Cox method, which we will refer to as \textit{Cox-MLP (CC)} and \textit{Cox-Time} respectively.
We will compare our methods to a classical linear Cox regression referred to as \textit{Classical Cox (Linear)}, DeepHit \citep{deephit}, and Random Survival Forests (RSF) \citep{Ishwaran2008}.
We will also compare to a proportional Cox method similar to DeepSurv \citep{DeepSurv}, but our version performs batched SGD by computing the negative partial log-likelihood in~\eqref{eq:cox_loss} on a subset of the data set. Furthermore, we choose not to restrict the network structure and optimization scheme to that of~\citet{DeepSurv}. Hence, this method is identical to our proportional Cox method in Section~\ref{sub:non_linear_cox}, except that it computes the negative partial log-likelihood of a batch, while we use case-control sampling in the loss function.
We will refer to these two methods as Cox-MLP (DeepSurv) and Cox-MLP (CC) respectively.
We do not compare with \citet{Luck2017} as their method is another proportional hazard method and is therefore restricted in all the same ways as our other proportional methods.
As we will show in Section~\ref{sub:results}, the proportional hazards assumption is very restrictive, and methods based on this assumption are therefore not able to compete with other methods such as DeepHit, RSF, and Cox-Time.

For the neural networks, we standardize the numerical covariates and encode the categorical covariates by entity embeddings~\citep{guo2016entity} half the size of the number of categories. For the Classical Cox (Linear) regression, we one-hot encoded the categorical covariates (dummy variables), and in RSF we simply passed the covariates without any transformations.
The networks are standard multi-layer perceptrons with the same number of nodes in every layer, ReLU activations, and batch normalization between layers. We used dropout, normalized decoupled weight decay \citep{adamwr}, and early stopping for regularization. SGD was performed by AdamWR~\citep{adamwr} with an initial cycle length of one epoch, and we double the cycle length after each cycle. Learning rates were found using the methods proposed by \citet{smith2017}.

As the four data sets are somewhat small, we scored our fitted models using 5-fold cross-validation, where the hyperparameter search was performed individually for each fold.
For all the neural networks, we performed a random hyperparameter search over 300 parameter configurations and chose the model with the best score on a held-out validation set.
We scored the proportional Cox methods by the partial log-likelihood, and for the Cox-Time method, we used the loss~\eqref{eq:loss_cox_time_first}. Model hyperparameter tuning for DeepHit and RSF was performed with the time-dependent concordance index \citep{Ctd}. However, we also include an RSF tuned in the manner proposed by the authors, i.e., by computing the concordance of the mortality~\citep[][Section 4.1]{Ishwaran2008}.
The two versions of RSF are in the following denoted RSF ($C^\text{td}$) and RSF (Mortality), respectively.
A list of the hyperparameter search spaces can be found in Appendix~\ref{sub:hyperparameter_tuning}.

\subsubsection{Results}
\label{sub:results}

We compare the methods using the metrics presented in Section~\ref{sec:evaluation_criteria}, i.e., the time-dependent concordance, the integrated Brier score, and the integrated binomial log-likelihood. While the concordance solely evaluates a method's discriminative performance, the Brier score and binomial log-likelihood also evaluate the calibration of the survival estimates.

Table~\ref{tab:concordance_surv} shows the time-dependent concordance, or $C^\text{td}$, averaged over the five cross-validation folds.
As expected, the methods that assume proportional hazards, in general, perform worse than the less restrictive methods, and Cox-MLP (DeepSurv) and Cox-MLP (CC) are very close to each other. 
We see that RSF ($C^\text{td}$) has better concordance than RSF (Mortality), which is expected as RSF ($C^\text{td}$) uses the same metric for hyperparameter tuning as for evaluation.
Cox-Time seems to do slightly better than the RSF methods, which is impressive as we have not used concordance for hyperparameter tuning.
DeepHit seems to have the best discriminative performance overall, but, as we will see next, this comes at the cost of poorly calibrated survival estimates.

\begin{table}[t]
    \centering
    \begin{tabular}{lrrrr}
        \toprule
        Method &         SUPPORT &        METABRIC &    Rot.\ \& GBSG &         FLCHAIN \\
        \midrule
        Classical Cox (Linear)       &           0.598 &           0.628 &           0.666 &           0.790 \\
        Cox-MLP (DeepSurv) &           0.611 &           0.636 &           0.674 &           0.790 \\
        Cox-MLP (CC)       &           0.613 &           0.643 &           0.669 &  \textbf{0.793} \\
        Cox-Time           &           0.629 &           0.662 &  \textbf{0.677} &           0.790 \\
        DeepHit            &  \textbf{0.642} &  \textbf{0.675} &           0.670 &           0.792 \\
        RSF (Mortality)    &           0.628 &           0.649 &           0.667 &           0.784 \\
        RSF ($C^\text{td}$)          &           0.634 &           0.652 &           0.669 &           0.786 \\
        \bottomrule
    \end{tabular}
    \caption{Concordance ($C^\text{td}$) for the experiments in Section~\ref{sub:five_small_data_sets}.}\label{tab:concordance_surv}
\end{table}

\begin{table}[t]
    \centering
    \begin{tabular}{lrrrr}
        \toprule
        Method &         SUPPORT &        METABRIC &    Rot.\ \& GBSG &         FLCHAIN \\
        \midrule
        Classical Cox (Linear)       &           0.217 &           0.183 &           0.180 &           0.096 \\
        Cox-MLP (DeepSurv) &           0.214 &           0.176 &           0.170 &  \textbf{0.092} \\
        Cox-MLP (CC)       &           0.213 &           0.174 &           0.171 &           0.093 \\
        Cox-Time           &           \textbf{0.212} &  \textbf{0.172} &  \textbf{0.169} &           0.102 \\
        DeepHit            &           0.223 &           0.184 &           0.178 &           0.120 \\
        RSF (Mortality)    &           0.215 &           0.175 &           0.171 &           0.093 \\
        RSF ($C^\text{td}$)          &  \textbf{0.212} &           0.176 &           0.171 &           0.093 \\
        \bottomrule
    \end{tabular}
    \caption{Integrated Brier score weighted by estimates of the censoring distribution for the experiments in Section~\ref{sub:five_small_data_sets}.}\label{tab:ibs_km}
\end{table}

\begin{table}[t]
    \centering
    \begin{tabular}{lrrrr}
        \toprule
        Method &          SUPPORT &         METABRIC &     Rot.\ \& GBSG &          FLCHAIN \\
        \midrule
        Classical Cox (Linear)       &           -0.623 &           -0.538 &           -0.529 &           -0.322 \\
        Cox-MLP (DeepSurv) &           -0.619 &           -0.532 &           -0.514 &  \textbf{-0.309} \\
        Cox-MLP (CC)       &           -0.615 &           -0.515 &           -0.509 &           -0.314 \\
        Cox-Time           &           -0.613 &  \textbf{-0.511} &  \textbf{-0.502} &           -0.432 \\
        DeepHit            &           -0.637 &           -0.539 &           -0.524 &           -0.487 \\
        RSF (Mortality)    &           -0.619 &           -0.515 &           -0.507 &           -0.311 \\
        RSF ($C^\text{td}$)          &  \textbf{-0.610} &           -0.517 &           -0.507 &           -0.311 \\
        \bottomrule
    \end{tabular}
    \caption{Integrated binomial log-likelihood weighted by estimates of the censoring distribution for the experiments in Section~\ref{sub:five_small_data_sets}.}\label{tab:imbll_km}
\end{table}

Tables~\ref{tab:ibs_km} and~\ref{tab:imbll_km} show the integrated Brier score and integrated binomial log-likelihood, both weighted with Kaplan-Meier estimates of the censoring distribution. Here, for both metrics, closer to zero is better.
We find that both metrics yield very similar results, as the orderings of the methods are almost identical.
First, we see that Cox-Time seems to generally perform the best, but it struggles with the FLCHAIN data set.
However, we note that, for FLCHAIN, Cox-MLP (DeepSurv) has the best integrated Brier score and binomial log-likelihood while Cox-MLP (CC) has the best concordance.
This indicates that the proportionality assumption is quite reasonable for this data set.

Again, we find that there is very little difference between Cox-MLP (DeepSurv) and Cox-MLP (CC), which is as expected.
The RSF methods generally perform well. Note that the two RSF methods perform equally well here, even though RSF ($C^\text{td}$) had the best concordance.
Classical Cox (Linear) does rather poorly, which was expected as it has very restrictive model assumptions.

Even though DeepHit, in general, had the best concordance, it has the worst integrated Brier score and binomial log-likelihood in three out of the four data sets.
The loss function in DeepHit, given by formula~\eqref{eq:loss_deephit} in Appendix~\ref{app:deephit}, is a convex combination of the negative log-likelihood and a ranking loss, determined by a parameter $\alpha$. For $\alpha = 1$ we obtain only the negative log-likelihood and for $\alpha=0$ we obtain only the ranking loss.
As we do hyperparameter tuning based on the concordance, we see that $\alpha$ tends towards smaller values, which results in excellent discriminative performance at the cost of poorly calibrated survival estimates.

\subsection{KKBox Churn Case Study}
\label{sec:kkbox_churn_prediction}

Thus far we have compared competing survival methodologies on fairly small data sets. We now perform a case study on a much larger data set, as this is more interesting in the context of neural networks.

The WSDM KKBox's churn prediction challenge was proposed in preparation for the 11th ACM International Conference on Web Search and Data Mining.\footnote{\url{https://www.kaggle.com/c/kkbox-churn-prediction-challenge}}
The competition was hosted by Kaggle in 2017, with the goal of predicting customer churn on a data set donated by KKBox, the leading music streaming service in Asia.
The available data provide us the opportunity to create a survival data set with event times and censoring indicators. 
We stick with the churn definition given by KKBox, were a customer is considered churned if he or she fails to resubscribe within 30 days after the previous subscription expired. 
Note, however, that our use of the data is not comparable to the Kaggle competition, as we work with survival times from the start of a subscription period, while they consider durations from a fixed calendar date.

KKBox provides multiple data sources, but as we are primarily interested in evaluating our methods, we spend less time on feature engineering and only use a subset of covariates with general customer information (e.g., city, age, price of subscription).
Furthermore, a customer that has previously churned and later resubscribed, is treated as a new customer with some covariate information describing the previous subscription history.
This gives us a total of 15 covariates. 
We split the data into a training, a testing, and a validation set, and some information about these subsets are listed in Table~\ref{tab:kkbox_dataset}.
A more in-depth description of the KKBox data set can be found in Appendix~\ref{app:kkbox_dataset}.

\begin{table}[t]
    \centering
    \begin{tabular}{lrrrrr}
        \toprule
        Data set &         Size  &  Churned   & Censored &  Prop. Censor  & Unique users \\
        \midrule                                                                          
        Train      &  1,786,333 &  1,279,358 &  506,975 &          0.28 &       1,582,202 \\
        Test       &   661,748  &   473489   &  188,259 &          0.28 &        586,001 \\
        Validation &   198,665  &   142,104  &   56,561 &          0.28 &        175,801 \\
        \bottomrule
    \end{tabular}
    \caption{Summary of the KKBox churn data set.}\label{tab:kkbox_dataset}
\end{table}

\subsubsection{Methods and Hyperparameter Tuning}

We use the same methods as in Section~\ref{sub:five_small_data_sets}. However, as this data set is very large, we replace the classical Cox regression with our Cox-SGD (Linear) method from Section~\ref{sub:cox_with_sgd}.

We standardize and encode the covariates in the same manner as for the smaller data sets. The networks and training are also the same as earlier, but we multiply the learning rate by 0.8 at the start of every cycle, as we found this to give more stable training.
The KKBox data set is quite large, so we are not able to explore as large a hyperparameter space as in the previous experiments.
Hence, we do not include weight decay, and perform a grid search over a small number of suitable parameters. 
The hyperparameter search is described in detail in Appendix~\ref{app:kkbox_model_specifications}.

The best configurations are given in Table~\ref{tab:model_architectures}.
For RSF, the hyperparameter search based on $C^\text{td}$ yielded 8 covariates sampled for each split and a minimum of 50 observations in each leaf node. With concordance of mortality as the validation criterion, the best fitted model used 2 covariates for splitting, and a minimum leaf node size of 10. Furthermore, we found that 500 trees were sufficient, as there was little improvement compared to 250 trees.

\begin{table}[t]
    \centering
    \begin{tabular}{lrrrrr}
        \toprule
        Method & Layers & Nodes & Dropout &  $\alpha$ & $\sigma$ \\
        \midrule
        Cox-MLP (DeepSurv) &        6 &     256 &     0.1 &    - &   - \\
        Cox-MLP (CC)       &        6 &     128 &       0 &    - &   - \\
        Cox-Time           &        8 &     256 &       0 &    - &   - \\
        DeepHit            &        6 &     512 &     0.1 &  0.001 &   0.5 \\
        \bottomrule
    \end{tabular}
    \caption{KKBox model architectures. $\alpha$ and $\sigma$ only applies the DeepHit (see Appendix~\ref{app:deephit}).}\label{tab:model_architectures}
\end{table}

\subsubsection{Results}

For evaluation, we fitted each of the methods five times and computed the time-dependent concordance index ($C^\text{td}$), the integrated Brier score (IBS), and the integrated binomial log-likelihood (IBLL) of the fitted models.
The median scores are presented in Table~\ref{tab:evaluation_survival}. We use the median because the two proportional Cox-MLP methods yielded rather unstable results, where some of the fitted models performed very badly.

\begin{table}[t]
    \centering
    \vspace{5mm}
    \begin{tabular}{lrrr}
        \toprule
        Method &    $C^\text{td}$ &    IBS &  IBLL \\
        \midrule
        Cox-SGD (Linear)   &  0.816          &  0.127          & -0.406 \\
        Cox-MLP (DeepSurv) &  0.841          &  0.111          & -0.349 \\
        Cox-MLP (CC)       &  0.844          &  0.119          & -0.379 \\
        Cox-Time           &  0.861          &  \textbf{0.107} & \textbf{-0.334} \\
        DeepHit            &  \textbf{0.888} &  0.147          & -0.489 \\
        RSF (Mortality)    &  0.855          &  0.112          & -0.352 \\
        RSF ($C^\text{td}$)          &  0.870          &  0.111          & -0.352 \\
        \bottomrule
    \end{tabular}
    \caption{Evaluation metrics for the KKBox data.}\label{tab:evaluation_survival}
\end{table}

From the table, we see that DeepHit continues to outperform the other methods in terms of concordance while having the worst performance in terms of IBS and IBLL\@.
Furthermore, Cox-Time has the best IBS and IBLL, while still providing a decent concordance.
RSF continues to do well across all metrics, while again, the tuning based on $C^\text{td}$ seems to yield better results than tuning base on the concordance of the mortality. 
Cox-SGD (Linear) does rather poorly, as it is very restricted, and serves more as a baseline in this context.
The Cox-MLP methods seem to again perform reasonably close to each other, at least when taking into account that we found both methods to yield rather unstable results. We are not sure why this was the case, but note that the combination of the flexible neural net and the proportionality constraint might be problematic for large data sets.

\begin{figure}[t]
    \begin{center}
        \vspace{5mm}
        \includegraphics[scale=0.45]{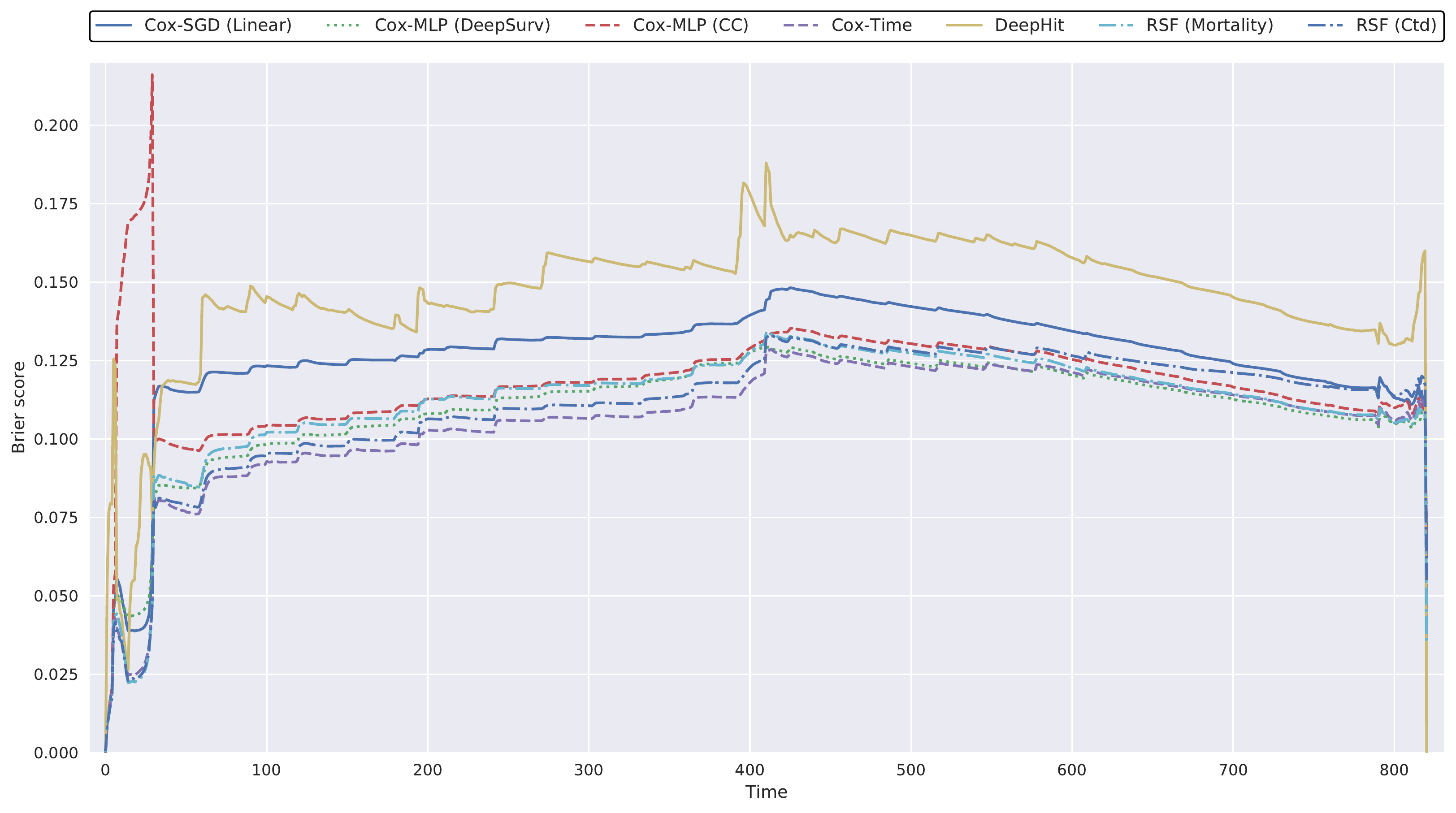}
    \end{center}
    \vspace{-7mm}
    \caption{Brier score on KKBox data set. The methods are the same as in Table~\ref{tab:evaluation_survival}.}\label{fig:kkbox_brier_score}
\end{figure}

In Figure~\ref{fig:kkbox_brier_score}, we display the Brier scores used to obtain the IBS\@. Again, we see that DeepHit does poorly for all durations. The instability of the Cox-MLP (CC) is also very apparent for shorter durations.
Cox-Time is clearly doing very well for all durations, but interestingly, we observe the Cox-MLP (DeepSurv) provides the best fitted model for larger durations.
We could make a corresponding figure for the binomial log-likelihood, but as it is very similar to the Brier score plot, and provide us with no new insights, we have not included it.

\subsubsection{Survival Curves}
\label{sub:kkbox_survival_curves}

One of the benefits survival models have over binary classifiers is their ability to produce survival curves.
In Figure~\ref{fig:kkbox_curves} in Appendix~\ref{app:kkbox_case_study}, we show nine examples of estimated survival curves from Cox-Time on the test data. 
The curves nicely illustrate the extent of detail the method has learned.

\begin{figure}[t!]
    \begin{center}
        \vspace{5mm}
        \includegraphics[scale=0.53]{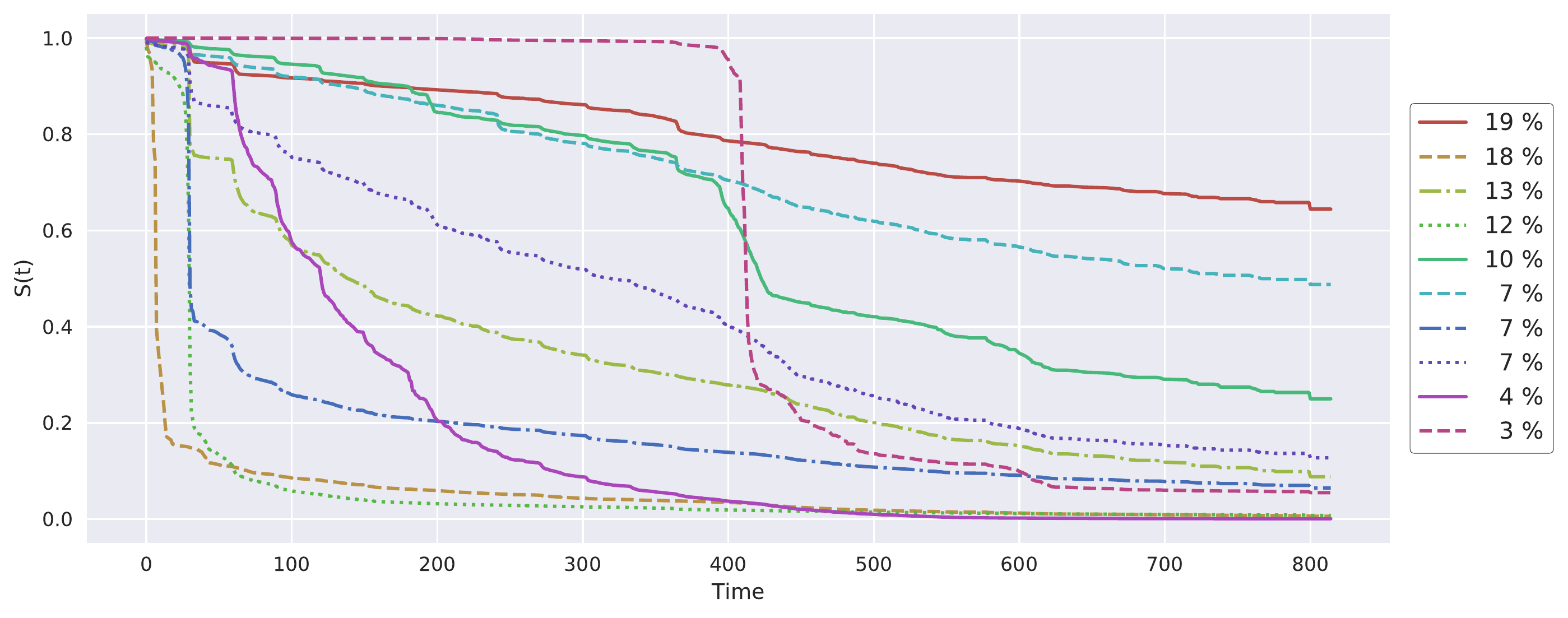}
    \end{center}
    \vspace{-7mm}
    \caption{Cluster centers of survival curves from KKBox test data. The centers were generated by K-means clustering on survival curves from Cox-Time. The legend gives the proportion of test data assigned to each cluster (rounded to nearest integer).}\label{fig:kkbox_kmeans}
\end{figure}

To obtain a more general view of the predictions, we cluster the estimated survival curves of the test set. 
For an equidistant grid $0 = \tau_0 < \tau_1 < \cdots < \tau_m$, the survival curve of individual $i$ is represented by a vector $[\hat S(\tau_0 \mid \x_i), \hat S(\tau_1 \mid \x_i), \ldots, \hat S(\tau_m \mid \x_i)]$, and by considering these as feature vectors, we apply K-means clustering to the test set with 10 clusters.
In Figure~\ref{fig:kkbox_kmeans}, we display the cluster centers and the proportions of the test set assigned to each of the clusters.
This is a reasonable approach for segmenting customers based on their churning behavior.
A natural next step would be to further investigate the clusters, but as we consider this somewhat outside the scope of this paper, we only make a few observations.
First, we  see that 19 \% of the customers are assigned to a cluster that does not provide much detailed information about their behavior, but instead provides a survival curve with a rather constant slope.
In sharp contrast, the second largest group (18 \%) is at high risk of churning immediately. 
Furthermore, we observe that many of the curves seem to have higher hazards at the end of each month (drops in the survival curves around 30, 60, 90 days), and we hypothesize that this is a result of customers paying for a full month at a time.

The smallest cluster, constituting only 3 \% of the test data, has a sharp drop around day 400.
Investigating the covariates of the assigned customers reveals that most of them had prepaid for a 411 days long subscription. However, the large drop after 400 days indicates that only around 25 \% of them were interested in continuing their subscription.

Our choice of 10 clusters is mainly motivated by how many curves that can be visualized in a single plot, without being too crowded.
Further increasing the number of clusters would likely reveal more detailed behavior.

\section{Discussion}
\label{sec:conclusion}

In this paper, we propose extensions of the Cox proportional hazards model.
By parameterizing the relative risk function of a Cox model with neural networks, we can model rich relationships between the covariates and event times.
Furthermore, we allow the networks to model interactions between the covariates and time, resulting in models that are no longer constrained by the proportionality assumption of the Cox model.
Building on methods for nested case-control studies, we propose a loss function that can be computed in batches, enabling the models to scale to large data sets.
We verify through simulation studies that fitting a Cox model using our proposed loss function gives results close to those obtained using the full Cox partial likelihood. 

We compare our suggested methodology with classical Cox regression, random survival forests (RSF), DeepHit, and DeepSurv on 5 real-world data sets, and find that our proposed Cox-Time method performs very well, and has the best overall performance in terms of integrated Brier score (IBS) and integrated binomial log-likelihood (IBLL).
DeepHit has, in general, the best discriminative performance, but this comes at the cost of poorly calibrated survival estimates.

Finally, we show how estimated survival curves (event probabilities as functions of time), can be used as a descriptive tool to better understand event-time data sets.
This is illustrated by an example where we cluster the survival estimates of a customer churn data set, and show that this customer segmentation provides a useful view of the churning process.

Interesting expansions of our methodology include the extension to handle multiple competing events, time-dependent covariates, dynamic predictions, and recurrent events.
Furthermore, it would be of interest to explore other data sources that require more advanced network structures such as convolutions and recurrent neural networks.
Finally, less computationally expensive alternatives for creating survival estimates for the Cox-Time method should be explored.


\acks{This work was supported by The Norwegian Research Council 237718 through the Big Insight Center for research-driven innovation.}



\appendix
\setcounter{equation}{0}
\setcounter{table}{0}
\setcounter{figure}{0}
\renewcommand{\theequation}{\thesection.\arabic{equation}}
\renewcommand{\theHequation}{\thesection.\arabic{equation}}
\renewcommand{\thetable}{\thesection.\arabic{table}}
\renewcommand{\theHtable}{\thesection.\arabic{table}}
\renewcommand{\thefigure}{\thesection.\arabic{figure}}
\renewcommand{\theHfigure}{\thesection.\arabic{figure}}

\section{Details on Hyperparameter Tuning}

In the following, we provide further details about the experiments conducted in Section~\ref{sec:experiments}.
Here we list hyperparameter configurations and details about the model fitting procedures.

\subsection{Four Common Survival Data Sets Tuning}
\label{sub:hyperparameter_tuning}

Table~\ref{tab:hyper_pars} gives the hyperparameter search space used for Rot.\ \& GBSG, SUPPORT, METABRIC, and FLCHAIN\@.
The square brackets describe continuous variables.
In the experiments in Section~\ref{sub:five_small_data_sets}, we sample 300 random parameter configurations for each method, for each fold of each data set.
In the table, 
``$\alpha$'' and ``$\sigma$'' are given in~\eqref{eq:loss_deephit} in Appendix~\ref{app:deephit}, ``Num.\ durations'' are the number of discrete durations (equidistant) used in DeepHit,
``$\lambda$'' is the penalty in~\eqref{eq:cox_penalty}, 
``Log duration'' refers to a log-transform of the durations passed to Cox-Time, 
``Ridge'' is a ridge penalty used in  classical Cox regression, 
``Split covariates'' and ``Size leaf'' are the number of covariates used for splitting, and the minimum node size of RSF\@.

\begin{table}[p]
    \centering
    \begin{tabular}{lr}
        \toprule
        Hyperparameter & Values\\
        \midrule
        Layers        &    \{1, 2, 4\} \\
        Nodes per layer & \{64, 128, 256, 512\} \\
        Dropout & [0, 0.7] \\
        Weigh decay & \{0.4, 0.2, 0.1, 0.05, 0.02, 0.01, 0\} \\
        Batch size & \{64, 128, 256, 512, 1024\} \\
        $\alpha$ (DeepHit) & [0, 1] \\
        $\sigma$ (DeepHit) & \{0.1, 0.25, 0.5, 1, 2.5, 5, 10, 100\} \\
        Num.\ durations (DeepHit) & \{50, 100, 200, 400\} \\
        $\lambda$ (Cox-Time and Cox-MLP (CC)) & \{0.1, 0.01, 0.001, 0\} \\
        Log durations (Cox-Time) & \{True, False\} \\ 
        Ridge (Cox (Linear)) & \{1000, 100, 10, 1, 0.1, 0.01, $10^{-3}$, $10^{-4}$, $10^{-5}$\} \\
        Split covariates (RSF) & \{2, 4, 6, 8\} \\
        Size leaf (RSF) & \{2, 8, 32, 64, 128\} \\
        \bottomrule
    \end{tabular}
    \caption{Hyperparameter search space for experiments on  Rot.~\& GBSG, SUPPORT, METABRIC, and FLCHAIN.}\label{tab:hyper_pars}
\end{table}

\subsection{KKBox Tuning}
\label{app:kkbox_model_specifications}

\begin{table}[p]
    \centering
    \begin{tabular}{lr}
        \toprule
        Hyperparameter & Values\\
        \midrule
        Layers        &    \{4, 6, 8\} \\
        Nodes per layer & \{128, 256, 512\} \\
        Dropout & \{0, 0.1, 0.5\} \\
        \midrule
        $\alpha$(*) & \{0, 0.001, 0.1, 0.2, 0.5, 0.8, 0.9, 0.99, 0.999, 1\} \\
        $\sigma$(*) & \{0.01, 0.1, 0.25, 0.5, 1, 10, 100\} \\
        Log durations(*) & \{True, False\} \\ 
        \midrule
        Split covariates & \{2, 4, 6, 8\} \\
        Size leaf & \{8, 10, 20, 50\} \\
        \bottomrule
    \end{tabular}
    \caption{KKBox hyperparameter configurations. (*) denotes parameters found with a two layer network with 128 nodes.}\label{tab:hyper_pars_kkbox}
\end{table}

Hyperparameters in the KKBox study were found by a grid search over the relevant parameters in Table~\ref{tab:hyper_pars_kkbox}.
The table consists of three sections, where the top represents the networks, the bottom represents RSF, and the middle contains network parameters that were found on a smaller network with two layers and 128 nodes.
``$\alpha$'' and ``$\sigma$'' controls the loss function of DeepHit, and we assumed it should generalize well across network structures. The same goes for whether or not we should log-transform the durations passed to Cox-Time.
Hence, to reduce the hyperparameter search, we found suitable values with a smaller network.

For the proposed Cox-MLP (CC) and Cox-Time, we used a fixed penalty $\lambda =0.001$ in~\eqref{eq:cox_penalty}.
All networks were trained with batch size of 1028, and the best performing architectures can be found in Table~\ref{tab:model_architectures}.

\setcounter{equation}{0}
\setcounter{table}{0}
\setcounter{figure}{0}
\section{Details from KKBox Churn Case Study}
\label{app:kkbox_case_study}

In the following, we provide some details of the KKBox case study that were exempt from the main article.

\begin{figure}[t]
    \begin{center}
        \vspace{5mm}
        \includegraphics[scale=0.6]{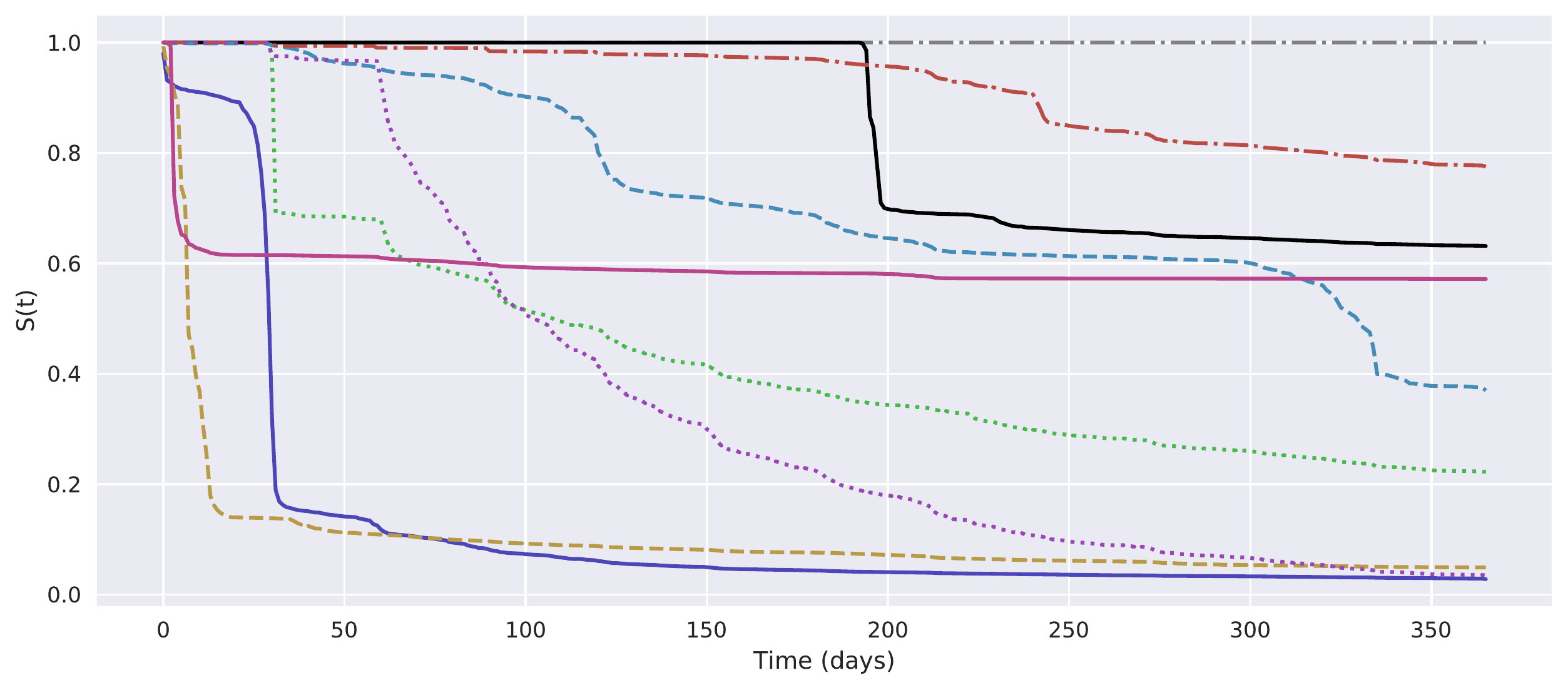}
    \end{center}
    \vspace{-7mm}
    \caption{Survival curves from the case study in Section~\ref{sec:kkbox_churn_prediction}, which models the times at which customers of a streaming service churn.
        Each curve gives the estimated probabilities for a customer not having churned (probabilities of still being a customer at any given time). The time axis shows the number of days since first subscription. The curves are generated by the Cox-Time method from Section~\ref{sub:non_proportional_cox}.}\label{fig:kkbox_curves}
\end{figure}

In Figure~\ref{fig:kkbox_curves}, we show an example of nine survival curves estimated by Cox-Time on the KKBox data set.
Each line represents an individual from the test set. It is clear that the Cox-Time method has learned to represent a variety of survival curves.

\subsection{KKBox Data Set}
\label{app:kkbox_dataset}

KKBox provides data consisting of general customer information (city, age, gender, initial registration), transaction logs listing how customers manage their subscription plans, and user logs containing some aggregate information of the customers' usage of the streaming service. 
KKBox defines a customer as churned if he or she has failed to resubscribe to their service more than 30 days after their last subscription expired. 

Through the transaction information, we can create a data set with survival times and censoring indicators. We keep KKBox's original churn definition of 30 days absence and calculate the survival time from the date of the first subscription or the earliest record of the customer in the transaction logs. Customers that have previously churned but resubscribed, are treated as new customers with some covariate information describing their previous subscription history. 

All covariates are calculated at the time of subscription of each customer, and they do not change with time. This is a simplification, as the covariates of a customer are typically not stationary.
However, as our objective is to evaluate the proposed methodology, we refrain from doing extensive feature engineering.
In this regard, we further disregard the user logs, as we get a reasonable set of covariates from the customer and transaction information.

The data sets are summarized in Table~\ref{tab:kkbox_dataset}. As some of the customers have churned multiple times, and are therefore included multiple times, the table also includes the number of unique customers in the respective data sets. 

We have a total of 15 covariates, where 7 are numeric and 8 are categorical.
The numerical covariates give the time between subscriptions (for customers that have previously churned), number of previous churns, time since first registration, number of days until the current subscription expires, listed price of the current subscription, the amount paid for the subscription, and age. All numerical covariates, except the number of previous churns, were log-transformed.
The categorical covariates are gender (3 categories), city (22 categories), how the customer registered (6 categories), and 5 indicator variables specifying if an individual has previously churned, if the subscription is canceled, if the age is unrealistic, if the subscription is automatically renewed, and if we do not know when the customer first registered.

\setcounter{equation}{0}
\setcounter{table}{0}
\setcounter{figure}{0}
\section{Additional Simulations}
\label{app:simulations}

In the following, we provide some additional simulations used to further investigate the proposed methods, and in Appendix~\ref{app:simulation_details} we explain how the simulated data were generated.
We again stress that the aim of the simulations is only to verify the expected behavior of our methods, and should not be interpreted as a general evaluation of them.

\subsection{Non-Linear Hazards}
\label{app:sim_non_linear_hazards}

Continuing from the simulations in Section~\ref{sec:simulations}, we do a simple study of the increased flexibility provided by replacing the linear predictor in a Cox regression by a neural network.
To evaluate this, we extend the simulations in Section~\ref{sec:simulations} by replacing $g(\x) = \boldsymbol \beta^T \x$ with the non-linear function 
\begin{align}
    \label{eq:sim_non_lin_ph}
    g(\x) = 
    \boldsymbol \beta^T \x +
    \frac{2}{3} (x_1^2 + x_3^2 + x_1 x_2 + x_1 x_3 + x_2 x_3),
\end{align}
where $x_i$ denotes element $i$ of $\x$. 
The simulations are otherwise unchanged from the linear case in Section~\ref{sec:simulations}.
We sample 10,000 training samples, 10,000 test samples, and 1,000 samples used for validation (validation data is used for early stopping),
and fit the Cox-SGD and classical Cox regression from Section~\ref{sec:simulations}.
Additionally, we fit the Cox model in Section~\ref{sub:non_linear_cox},
where $g(\x)$ is parameterized by a one-hidden layer MLP (multilayer perceptron) with 64 hidden nodes and ReLU activations. 
This model will be referred to as \textit{Cox-MLP} (we drop (CC) as we do not consider DeepSurv here).

\begin{figure}[t!]
    \begin{center}
        \vspace{5mm}
        \includegraphics[scale=0.5]{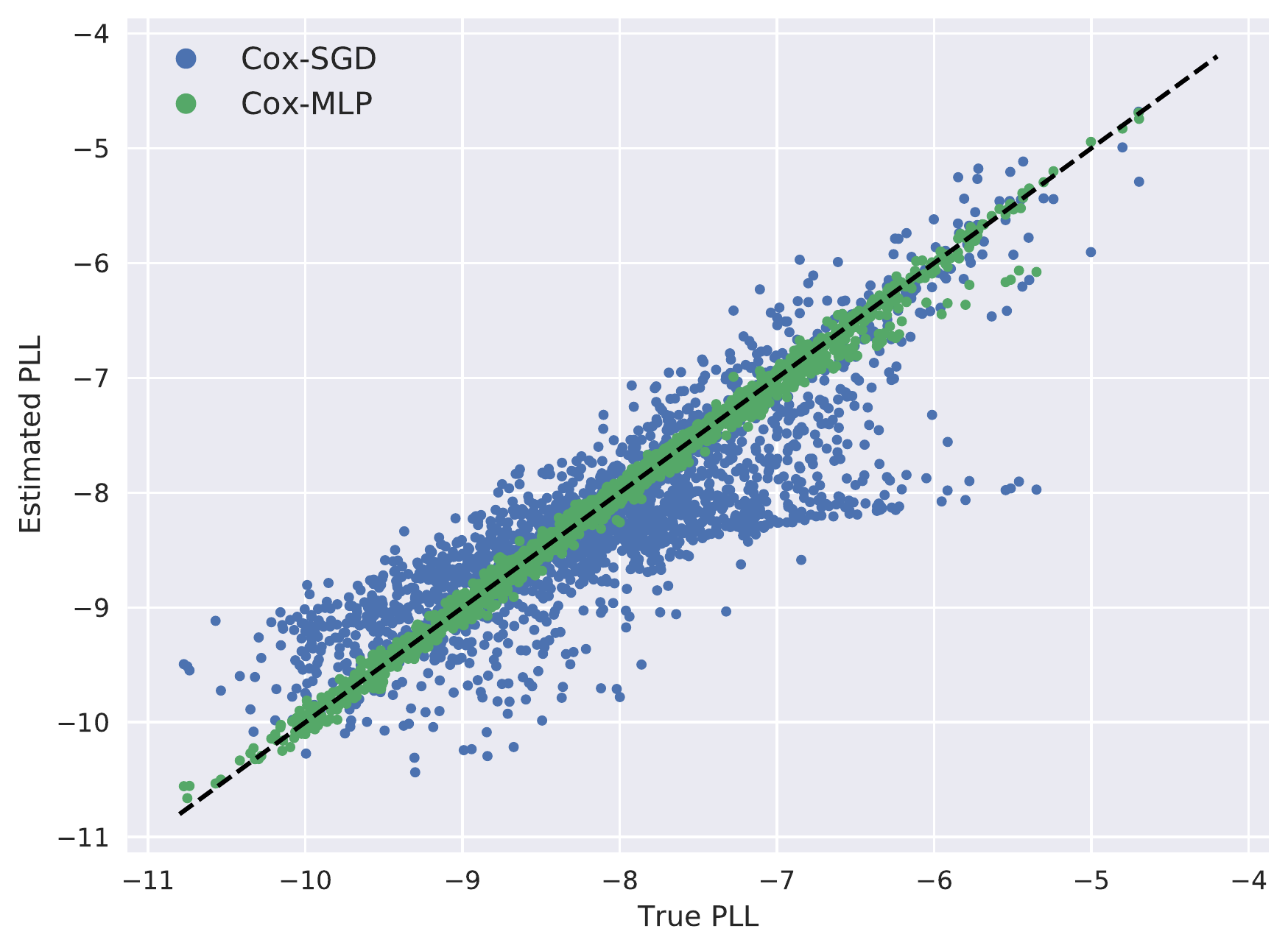}
    \end{center}
    \vspace{-7mm}
    \caption{Partial log-likelihood estimates of Cox-SGD  and Cox-MLP plotted against the true partial log-likelihoods, for 2,000 samples of the test set.}\label{fig:pll_non_linear}
\end{figure}

In Figure~\ref{fig:pll_non_linear} we have, for 2,000 individuals of the test set, plotted the individual partial log-likelihood estimates of Cox-SGD and Cox-MLP against the true individual partial log-likelihoods. That is, for each $i$ with $D_i = 1$, we plot 
\begin{align*}
    \hat \ell_i = - \log \left( \sum_{j \in \mathcal R_i} \exp[\hat g(\x_j) - \hat g(\x_i)] \right)
\end{align*}
against the true $\ell_i$.
The closer a method estimates $g(\x)$ to the true predictor in~\eqref{eq:sim_non_lin_ph}, the closer the scatter plot should be to the identity function (straight line with slope 1 through the origin).
As expected, we see that Cox-MLP produces likelihood estimates very close to the true likelihoods, while Cox-SGD struggles to represent this non-linear function.

\subsection{Non-Proportional Hazards}
\label{app:sim_non_proportional_hazards}

In our final simulations, we investigate the effect of removing the proportionality constraint in the Cox models.
Building on the previous simulations, we add a time dependent term to the risk function. The hazard function is now given by $h(t \mid \x) = h_0 \exp[g(t, \x)]$, with $h_0 = 0.02$ and
\begin{align*}
    g(t, \x) &= a(\x) + b(\x) \; t, \\
    a(\x) &= g_{ph}(\x) + \text{sign}(x_3), \\
    b(\x) &= | 0.2\,  (x_0 + x_1) + 0.5 \, x_0  x_1 |,
\end{align*}
were $g_{ph}(\x)$ is the function in~\eqref{eq:sim_non_lin_ph}. 
The simulations are otherwise unchanged from the previous experiments.
We require the term $b(\x)$ to be non-negative to ensure that the hazards increase with time.  
Furthermore, we have added $\text{sign}(x_3)$ to $a(x)$ as this is essentially equivalent to having two different baselines, $\tilde h_0 = h_0 \exp[\text{sign}(x_3)] \in \{0.0074, 0.054\}$. 
Both of these choices are motivated by our attempt to produce reasonable looking survival curves.

We sample 10,000 training samples, 10,000 test samples, and 2,000 samples for validation, and fit a Cox-MLP model and a Cox-Time model.
Both Cox-MLP and Cox-Time parameterize $g$ with a 4 hidden layer MLP with ReLU activations, 128 nodes in each layer, and dropout between layers with rate 0.1. 

\begin{figure}[t]
    \begin{center}
        \vspace{5mm}
        \includegraphics[scale=0.5]{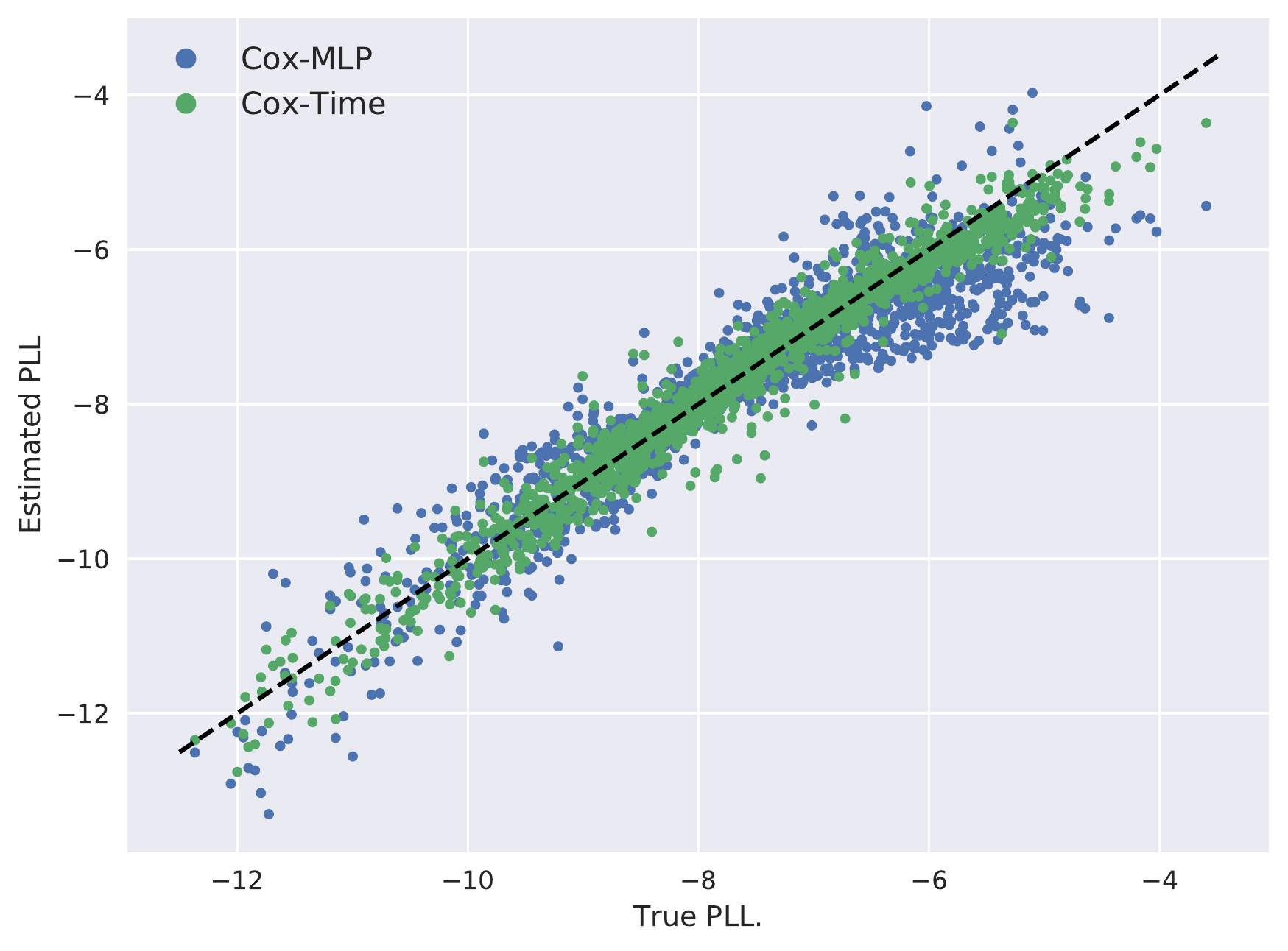}
    \end{center}
    \vspace{-7mm}
    \caption{Partial log-likelihood estimates of Cox-MLP  and Cox-Time plotted against the true partial log-likelihoods, for 2,000 samples of the test set.}\label{fig:pll_non_prop}
\end{figure}

\begin{figure}[t]
    \begin{center}
        \vspace{5mm}
        \includegraphics[scale=0.54]{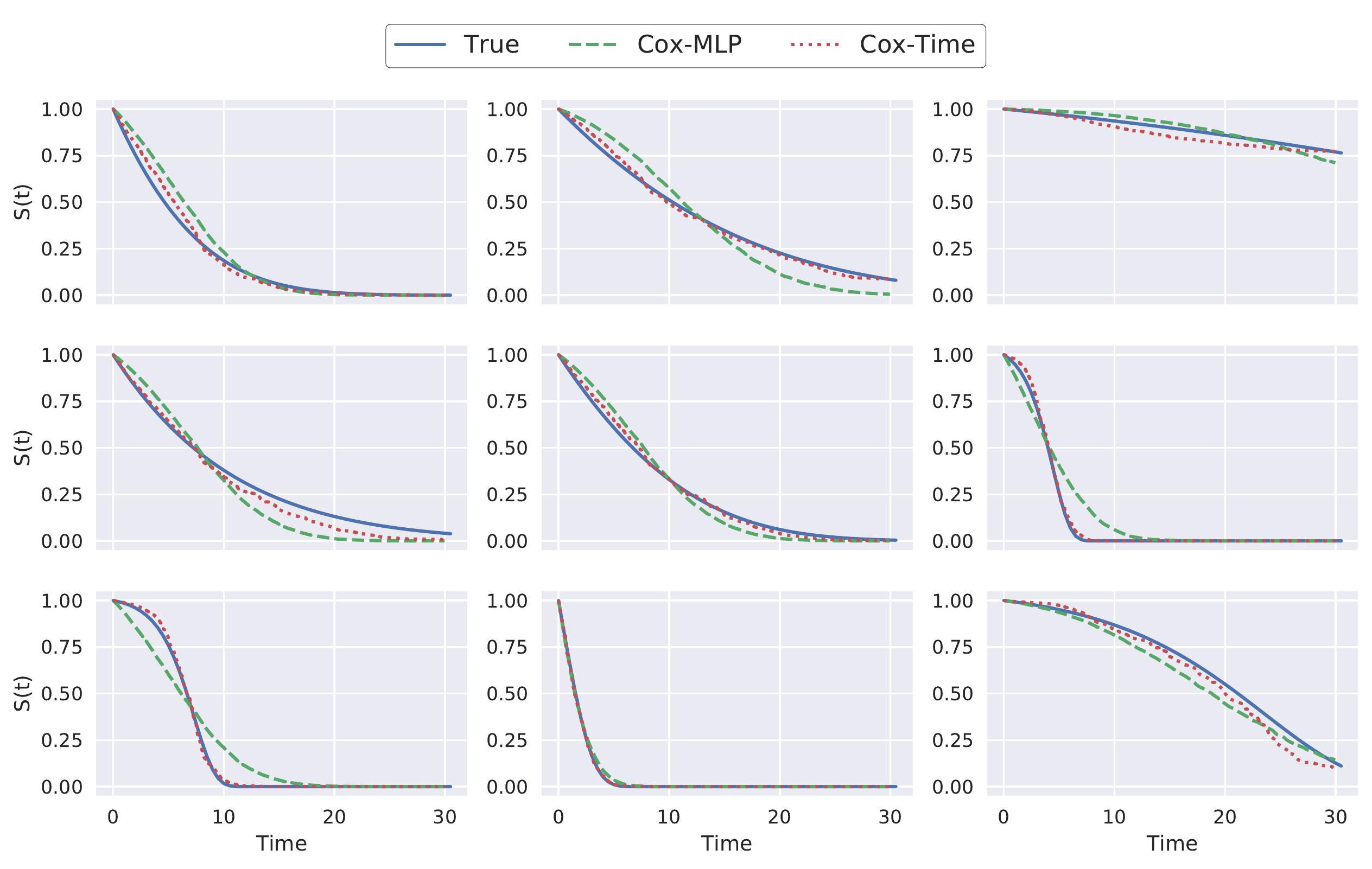}
    \end{center}
    \vspace{-7mm}
    \caption{Examples of survival curves $\hat S(t \mid \x)$ from the test set in the non-proportional simulation study.}\label{fig:survival_curves_non_prop}
\end{figure}

In Figure~\ref{fig:pll_non_prop}, we have plotted the estimated partial log-likelihoods of Cox-MLP and Cox-Time against the true partial log-likelihoods for 2,000 samples of the test set. 
Though the difference between the methods is not very large, Cox-Time appears to capture the true values better than Cox-MLP\@.

To further illustrate some of the differences between Cox-MLP and Cox-Time, we have in Figure~\ref{fig:survival_curves_non_prop} plotted nine survival curves $\hat S(t \mid \x)$ from the test set. The figure shows both the true curves and the curves estimated by the two methods.
It is seen that Cox-Time is able to estimate the true survival curves better than those produced by Cox-MLP\@.

\subsection{Simulation Details}
\label{app:simulation_details}

In the following, we explain in detail how we generated our simulated data sets in Section~\ref{sec:simulations} and Appendices~\ref{app:sim_non_linear_hazards} and~\ref{app:sim_non_proportional_hazards}.
We want to generate survival times $\Ts$ from relative risk models of the form
\begin{align}
    \label{eq:sim_const}
    h(t \mid \x) = h_0(t) \exp[g(t, \x)].
\end{align}
If the function $g$ does not depend on $t$, we have a proportional hazards model, which is just a special case of the relative risk models.
Let $H(t \mid \x)$ denote the continuous (increasing) cumulative hazard 
$H(t \mid \x) = \int_0^t h(s \mid \x)\, ds$, and let $V$ be exponentially distributed with parameter 1.
Then we can obtain survival times through the inverse cumulative hazard
$\Ts = H^{-1}(V \mid \x)$.
This can be shown to hold through the hazard's relationship to the survival function in~\eqref{eq:survival_from_hazard},
\begin{align*}
    S(t \mid \x) = \Pm(\Ts > t \mid \x) = \Pm(H^{-1}(V \mid \x) > t) = \Pm(V > H(t \mid \x)) = \exp[-H(t \mid \x)], 
\end{align*}
as $V$ is exponentially distributed with $\Pm(V > v) = \exp(-v)$.
Hence, we can obtain survival times by transforming generated samples from an exponential distribution.

To obtain an analytical expression for the inverse cumulative hazard $H^{-1}(v \mid \x)$, we restrict the models in~\eqref{eq:sim_const} to have constant baseline $h_0$, in addition to a simple time dependence in $g(t, \x)$.
For the linear predictor in Section~\ref{sec:simulations} and the non-linear predictor in Appendix~\ref{app:sim_non_linear_hazards} we simulate with proportional hazards, meaning $g(t, \x) = g(\x)$. Hence, we get cumulative hazards and inverse cumulative hazards of the form
\begin{align*}
    H(t \mid \x) &= t\, h_0 \exp[g(\x)],\\
    H^{-1}(v \mid \x) &= \frac{v}{h_0 \exp[g(\x)]}.
\end{align*}

In the non-proportional simulations in Appendix~\ref{app:sim_non_proportional_hazards}, we add a time dependent term $g(t, \x) = a(\x) + b(\x) \; t$, which gives us
\begin{align*}
    H(t \mid \x) &= \frac{h_0 \exp[a(\x)]\, (\exp[b(\x) \, t] - 1)}{b(\x)},\\
    H^{-1}(v \mid \x) &= \frac{1}{b(\x)} 
    \log \left (1 +  \frac{v\, b(\x)}{h_0\, \exp[a(\x)]} \right ).
\end{align*}

\setcounter{equation}{0}
\setcounter{table}{0}
\setcounter{figure}{0}
\section{DeepHit}
\label{app:deephit}

DeepHit by \citet{deephit} is a discrete survival model for competing risks. However, as we only consider one type of event, we will express the method in terms of a single cause.
DeepHit considers time to be discrete, so to fit it to the continuous-time data sets in Section~\ref{sec:experiments}, we discretize the event times with an equidistant grid between the smallest and largest duration in the training set.
The number of discrete time-points is considered a hyperparameter, given by ``Num.\ durations'' in Table~\ref{tab:hyper_pars}.

Now, assume time is discrete with $0 = \tau_0 < \tau_1 < \tau_2 < \cdots < \tau_m$.
Let $\mathbf{y}(\x) = {[y_0(\x), y_1(\x), \ldots, y_m(\x)]}^T$ denote the output of a neural net with covariates $\x$ and a softmax output activation. Then, $\mathbf{y}(\x)$ can be interpreted as the estimated probability mass function of the duration times
$y_j(\x_i) = \hat \Pm(\Ts_i = \tau_j \mid \x_i)$.
The estimated survival function is then given by 
\begin{align*}
    \hat S(\tau_j \mid \x) = 1 - \sum_{k=1}^j y_k(\x),
\end{align*}
and the discrete negative log-likelihood corresponding to the continuous version in~\eqref{eq:full_likelihood}, is
\begin{align*}
    \text{loss}_{L} = - \sum_{i=1}^N \left[ D_i \log (y_{e_i} (\x_i)) + (1 - D_i) \log(\hat S[\tT_i \mid x_i]) \right].
\end{align*}
Here, $e_i$ denotes the index of the event time for observation $i$, i.e., $\tT_i = \tau_{e_i}$.
Furthermore, DeepHit adds a second loss that attempts to improve its ranking capabilities,
\begin{align}
    \label{eq:loss_deephit_rank}
    \text{loss}_{\text{rank}} = \sum_{i,j} D_i\, \mathbbm{1}\{\tT_i < \tT_j\} \exp \left( \frac{\hat S(\tT_i \mid \x_i) - \hat S(\tT_i \mid \x_j)}{\sigma} \right).
\end{align}
The loss of DeepHit is a combination of these two losses, where the ranking loss is scaled by a constant. We deviate slightly from the original implementation here and instead use a convex combination of the two,
\begin{align}
    \label{eq:loss_deephit}
    \text{loss} = \alpha\,  \text{loss}_{L} + (1-\alpha)\, \text{loss}_{\text{rank}},
\end{align}
where $\alpha$ and $\sigma$ from~\eqref{eq:loss_deephit_rank} are considered hyperparameters.



\vskip 0.2in
\bibliography{bibliography}

\end{document}